\documentclass[12pt,english]{article}
\usepackage[english]{babel} % BibTex
\usepackage[nottoc]{tocbibind} % Include References
\usepackage[utf8]{inputenc}
\usepackage{amsmath, amssymb}
\usepackage{commath}
\usepackage{graphicx}
\usepackage{textcomp}
\usepackage{comment}
\usepackage{caption}
\usepackage{subcaption}
\usepackage[toc,page]{appendix}
\usepackage[a4paper,bindingoffset=0.3in,left=0.4in,right=0.6in,top=0.5in,bottom=0.5in,footskip=.25in]{geometry}
\usepackage{xcolor}

\usepackage{tikz}
\usetikzlibrary{positioning, fit, arrows.meta, shapes}
\usepackage{url}
\usepackage{chngcntr}
\usepackage{hyperref}

\newcommand{\empt}[2]{$#1_{#2}$}
\usepackage{float}

\title{Towards Robust Car Following Dynamics Modeling via Blackbox Models: Methodology, Analysis, and Recommendations}
\author{Muhammad Bilal Shahid \thanks{\href{mailto://belal@iastate.edu}{belal@iastate.edu}} \footnote{Corresponding author} \and Cody Fleming \thanks{\href{mailto://flemingc@iastate.edu}{flemingc@iastate.edu}}} 

\date{Department of Mechanical Engineering\\ Iowa State University\\ Ames, IA 50011}

\begin{document}

\maketitle

\section*{Abstract}\label{sec:abst}
The selection of the target variable is important while learning parameters of the classical car following models like GIPPS, IDM, etc. There is a vast body of literature on which target variable is optimal for classical car following models, but there is no study that empirically evaluates the selection of optimal target variables for black-box models, such as LSTM, etc. The black-box models, like LSTM and Gaussian Process (GP) are increasingly being used to model car following behavior without wise selection of target variables. The current work tests different target variables, like acceleration, velocity, and headway, for three black-box models, i.e., GP, LSTM, and Kernel Ridge Regression. These models have different objective functions and work in different vector spaces, e.g., GP works in function space, and LSTM works in parameter space. The experiments show that the optimal target variable recommendations for black-box models differ from classical car following models depending on the objective function and the vector space. It is worth mentioning that models and datasets used during evaluation are diverse in nature: the datasets contained both automated and human-driven vehicle trajectories; the black-box models belong to both parametric and non-parametric classes of models. This diversity is important during the analysis of variance, wherein we try to find the interaction between datasets, models, and target variables. It is shown that the models and target variables interact and recommended target variables don't depend on the dataset under consideration. \\

\textbf{Keywords}: car following models; target variable selection; LSTM car following; Gaussian Process car following; GIPPS; Intelligent driver model; full velocity difference model 

\section{Introduction}\label{sec:intro}
For intelligent transportation systems to reach their full potential, several challenges must be addressed. Predicting the intent and future behavior of neighboring vehicles (and other objects) is paramount for achieving safe and efficient trajectory planning and control as well as for predicting traffic flows and capacities, which enables intelligent system-level interventions for improving traffic safety and throughput.

Car following (CF) models provide a means for predicting vehicle behavior, and potentially larger traffic phenomena by modeling behavior and dynamics in leader-follower scenarios. Various car following models have been formulated to represent how a vehicle reacts to the changes in the relative positions and/or relative velocities of the vehicle ahead. There are myriad CF models, from symbolic ``first principles'' models to, more recently, purely data-driven models.

All of these models must be calibrated to real and/or simulated data; that is, the models' parameters must be updated such that the model predictions best fit the data. This gives rise to questions about how to perform this calibration, including:

\begin{enumerate}%[noitemsep]
    \item what kind of model structure to use, 
    \item what data to use to tune the parameters, 
    \item which variables (e.g. headway, velocity, acceleration) -- i.e. {\em target variable} -- should be used for this optimization?
\end{enumerate}
%\\

This last point is subtle and requires further explanation by way of example. To calibrate a CF model, one might try to minimize the mean-squared error on displacement, but this displacement is found by numerically integrated acceleration, which is the output of the model. In this case, displacement is the so-called target variable, while acceleration is the predicted variable. One could have instead chosen to minimize mean-squared error directly on acceleration; whereby the target and predicted variables are the same. 

This paper demonstrates that these choices -- i.e., how one chooses to address the questions above -- have significant impact on CF model performance. The dynamics of CF are modeled using three black-box approaches: Gaussian Processes (GP), Kernel-Ridge, and LSTM. It is empirically shown that the choices of target variables depend on the objective function and vector space of the model under consideration. After comparing target variable selection in black-box approaches versus classical approaches, we then show that the recommendations for target variable selection are robust to data that is qualitatively different. %The optimal choices of target variables for black-box and classical models are compared. Three datasets are used: ASTAZERO, which contains automated vehicle trajectories, and NAPOLI and JIANG, which contain trajectories from human-driven vehicles.  The analysis of variance is conducted to show that the recommendations of target variables for different models are robust to the dataset under consideration.\\
The outline of the paper is as follows: section \ref{sec:related} includes the literature review, section \ref{sec:models} discusses the different classical CF models, GP, Kernel-Ridge, and LSTM briefly; section \ref{sec:resultss} presents the results of the study; section \ref{sec:conclusion} reports the conclusions.

\section{Related Work}\label{sec:related}
Developing and improving CF models is an active area in traffic flow theory with two overarching topics. The first involves classical CF models (e.g. Gipps, Wiedemann, etc.); the primary focus area includes the selection of various algorithms to find the optimal parameters for these models for fitting to real or simulated data. The second area involves using machine learning algorithms instead of the classical models. 

The most popular classical CF models include Gipps, IDM, Wiedemann, Full Velocity Difference Model (FVDM), and Optimal Velocity (OV). The common theme in these models is that the parameters must be tuned to yield good predictive capability. These models have varying performance contingent upon the Goodness of fit (GoF) and Measure of Performance (MoP) used to find the optimal parameters. Note that we use the term {\em target variable} instead of MoP in later sections. In \cite{pourabdollah2017calibration}, the authors optimize three CF models (IDM, Wiedemann, and Krauss) with a novel GoF, based on power demand, which depends on all three MoPs (acceleration, velocity, and headway); IDM performed better than the rest of the models. When a CF model is calibrated using an optimization algorithm, it typically has good performance on the calibrated dataset but this may not necessarily result in good performance on a wide range of traffic conditions. %; that is, it does not have a good transferability. 
To that end, Vasconcelos et al. \cite{vasconcelos2014calibration} introduced sequential calibration wherein the parameters are determined one-by-one (individually) based on their respective physical meanings; the results were compared with an approach where all the parameters are simultaneously determined. The GoF and MoP were RMSE and s, respectively. To make the estimated parameters applicable to a wide variety of traffic situations, another effort was made by Markou \cite{markou2019dynamic}. In this paper, a distribution of parameters was estimated instead of just a point estimate of a parameter value; the idea of having a distribution over parameters is to capture a wide range of traffic conditions. The GoF used was Normalized root mean square error (NRMSE). 

So-called data-driven\footnote{We should note that classical models are also driven by data, but the reliance on data is different by degree from, say, machine learning techniques such as deep neural networks or Gaussian Processes. The former rely on physical insight into the problem to create parametric models while the latter have general, abstract architectures that do not (typically or necessarily) rely on specific physical insights into the problem.} methods rely on data to infer the underlying dynamics of the system. Wang et al. \cite{wang2021personalized} used the Gaussian Processes (GP) regression model to learn the CF model based on both synthetic data (generated using IDM) and naturalistic driving data (collected using Logitech racing wheel and Unity); the GP model was optimized on acceleration and it outperformed classical CF models such as IDM and OVM. The performance of these methods is also impacted by the number of independent variables used in the problem formulation. The typical choice of independent variables include follower's velocity ($v$), leader's velocity ($v_{l}$), space headway (s), and difference of leader and follower's velocity ($\Delta v$). One of the works by Soldevila \cite{soldevila2021car} focuses on the choice of independent variables. They included the status of traffic light as an independent variable and concluded that it impacts the driver's acceleration. The GP method requires training even though it is considered as a non-parametric method. A non-parametric method that does not need to be trained and predict the output based on a few similar cases from the training data is k-nearest neighbor regression (kNN). This method---despite being parsimonious---is able to replicate the fundamental diagrams as well as the periodic traffic oscillations \cite{he2015simple}.

There are at least two works that studied the impact of different combinations of GoFs and MoPs for classical CF models. In \cite{punzo2012can}, the author focuses on various possible combinations of different types of algorithms (Downhill simplex, Genetic Algorithm, and OptQuest Multistart), different MoPs (v and s), and different GoFs (Root mean square error (RMSE), Mean absolute error, GEH statitic, and Theil's inequality coefficient) to identify the best performing combination. Another paper rigorously addresses the impact of different combinations of GoFs and MoPs on the performance of CF model \cite{punzo2021calibration} -- this paper considered 7 types of GoFs and 4 types of MoPs and also reviews other work that implement different combinations of GoF and MoP. 

Even though there are many papers on both classical and data-driven CF models, the impact of different combinations of \textit{datasets}, \textit{target variables}, and \textit{model type} on predictions across three variables ($a$, $v$, and $s$) is not addressed well in the literature. Recent studies in the classical CF model literature have attempted to isolate the impact of target variable selection. To the best of our knowledge, no work has assessed the impact of dataset, target variable, and model selection in a rigorous way for {\em data-driven CF models} and then comparing the outcomes with those for classical CF models. This paper is the first to do such a quantitative and qualitative evaluation.

\section{Models}\label{sec:models}

This section provides a brief background on the underlying mathematical formalism of several types of models. Starting with the classical car following models, the formulations of three learning algorithms (Gaussian Process, kernel-Ridge, and Long Short Term Memory networks) are then described briefly.
\subsection{Classical CF Models}
There are four types of classical CF models considered: IDM, Gipps, FVDM-CTH, and FVDM-SIGMOID. The symbolic forms of IDM, Gipps, and FVDM are given in Eqs. (1), (2), and (3), respectively. \\
\begin{equation} \label{eq:1}
    a_{n}(t) = \boldsymbol{a_{max.n}}.\left[1-\left(\frac{v_{n}(t)}{\boldsymbol{V_{max,n}}}\right)^{\boldsymbol{\delta_{n}}}-\left(\frac{\boldsymbol{s_{0,n}}+\boldsymbol{T_{n}}.v_{n}(t)-\frac{v_{n}(t).\Delta v_{n}(t)}{2.\sqrt{\boldsymbol{a_{max,n}.b_{n}}}}}{s_{n}(t)}\right)^{2}\right]
\end{equation}
\begin{align} \label{eq:2}
    a_{n}(t) &= \frac{min \begin{cases}
    v_{n}(t)+2.5.\boldsymbol{a_{max,n}.\tau_{n}}.\left(1-\frac{v_{n}(t)}{\boldsymbol{V_{max,n}}}\right).\sqrt{0.025+\frac{v_{n}(t)}{\boldsymbol{V_{max,n}}}}, &\\
    -b_{n}\left(\frac{\boldsymbol{\tau_{n}}}{2}+\boldsymbol{\theta_{n}}\right)+\sqrt{\boldsymbol{b_{n}^{2}}.\left(\frac{\boldsymbol{\tau_{n}}}{2}+\boldsymbol{\theta_{n}}\right)^{2}+\boldsymbol{b_{n}}\left[2.(s_{n}(t)-\boldsymbol{s_{0,n}})-\boldsymbol{\tau_{n}}.v_{n}(t)+\frac{v_{n-1}(t)^{2}}{\boldsymbol{\hat{b_{n}}}}\right]}
    \end{cases}}{\boldsymbol{\tau_{n}}}-\frac{v_{n}(t)}{\boldsymbol{\tau_{n}}}
\end{align}
\begin{equation} \label{eq:3}
    a_{n}(t) = \boldsymbol{K_{1,n}}.\left[V(s_{n}(t))-v_{n}(t)\right]+\boldsymbol{K_{2,n}}.\Delta v_{n}(t)
\end{equation}\\
The $V(S_{n}(t))$ in FVDM model (Eq. 3) has two different forms given below in Eqs. 4 and 5; these two froms of $V(S_{n}(t))$ correspond to two variants of FVDM called FVDM-CTH and FVMD-SIGMOID, respectively \cite{di2014distributed,di2019cooperative,zhang2017hierarchical}. 
\begin{equation} \label{eq:4}
    V(s_{n}) = \begin{cases}0 & ifs_{n} \leq \boldsymbol{s_{0,n}}\\
                \frac{s_{n}-\boldsymbol{s_{0,n}}}{\boldsymbol{T_{n}}} & if s_{n} \leq \boldsymbol{s_{0,n}} + \boldsymbol{T_{n}.V_{max,n}}\\
                \boldsymbol{V_{max,n}} & elsewhere
                \end{cases}
\end{equation}
\begin{equation} \label{eq:5}
    V(s_{n}) = \begin{cases}0 & if s_{n} \leq \boldsymbol{s_{0,n}}\\
                \frac{\boldsymbol{V_{max,n}}}{2}\left[1-\cos{\left(\pi \frac{s_n-\boldsymbol{s_{0,n}}}{\boldsymbol{T_{n}}.\boldsymbol{V_{max,n}}}\right)}\right] & if s_{n} \leq \boldsymbol{s_{0,n}} + \boldsymbol{T_{n}}.\boldsymbol{V_{max,n}}\\
                \boldsymbol{V_{max,n}} & elsewhere
                \end{cases}
\end{equation}

In Eqs.~\ref{eq:1},~\ref{eq:2},~\ref{eq:3},~\ref{eq:4}, and~\ref{eq:5}, the following vehicle's position, velocity and acceleration are denoted by $x_{n}(t)$, $v_{n}(t)$, and $a_{n}(t)$, respectively; the inter-vehicle distance and speed difference from the leading vehicle are given as $\Delta x_{n}$ and $\Delta v_{n}(t)$, respectively; the leading vehicle length is indicated by $l_{n-1}$; and the net inter-vehicle spacing is $s_{n}(t) = \Delta x_{n} - l_{n-1}$. The parameters of each model are indicated by bold letters.

An integration method with a time-step of 0.1s was used to obtain vehicle speed and position at next time-step as shown in Eqs. (6) and (7) \cite{treiber2006delays, treiber2015comparing}.
\begin{equation}
    v_{n}(t+\Delta t) = v_{n}(t) + a_{n}(t).\Delta t 
\end{equation}
\begin{equation}
    x_{n}(t+\Delta t) = x_{n}(t) + \frac{v_{n}(t+\Delta t) + v_{n}(t)}{2}.\Delta t
\end{equation}

%Now the mathematical forms of three learning algorithms (GP, kernel-Ridge, and LSTM) used will be described briefly.

\subsection{Gaussian Processes Regression}

A Gaussian Process (GP) assumes that the training data $\mathcal{D}=\{(\boldsymbol{x_{n}},y_{n}) : n= 1: N\}$ comes from a noisy process as shown in Eq.~\ref{eq:8}; where $\boldsymbol{x_{n}} \in \mathbb{R}^{d}$, $y_{n} \in \mathbb{R}$, and $\epsilon_{n} \sim \mathcal{N}(0, \sigma_{n}^2)$. 
\begin{equation} \label{eq:8}
    y_{n} = f(\boldsymbol{x}_{n}) + \epsilon_{n}
\end{equation}
For $N$ training points, the $N \times d$ feature matrix and $N \times 1$ output vector can be denoted by $\mathbf{X}$ and $\boldsymbol{y}$, respectively. The mean and covariance of the observed noisy responses denoted by $\mu_{X}$ and $Cov[y|\mathbf{X}]$ can be written as: 
\begin{align}
    \mu_{X} &= (m(\boldsymbol{x}_{1}),......,m(\boldsymbol{x}_{n})) \label{eq:9}\\
    Cov[y|\mathbf{X}] &= \mathbf{K}_{X, X} + \sigma_{y}^{2}\mathbf{I}_{N} \label{eq:10}
\end{align}
where $\mathbf{K}_{X, X}$ is kernel matrix of size $N \times N$ with each entry $\mathcal{K}(\boldsymbol{x}_{i}, \boldsymbol{x}_{j}$) being a similarity measure between training inputs $\boldsymbol{x}_{i}$ and $\boldsymbol{x}_{j}$. Now, given a test set $\mathbf{X_{*}}$ with a size of $N_{*} \times D$, the task is to predict the function outputs $\boldsymbol{f}_{*} = [f(\boldsymbol{x}_{1}),.....,f(\boldsymbol{x}_{N_{*}})]$. The joint density of the observed data and test data is given by the following equation.
\begin{equation}
    \begin{pmatrix}\boldsymbol{y}\\ \boldsymbol{f}_{*} \end{pmatrix} \sim \mathcal{N}\left(\begin{pmatrix}
    \boldsymbol{\mu}_{X} \\
    \boldsymbol{\mu_{*}}
    \end{pmatrix},
    \begin{pmatrix}
    \mathbf{K}_{X,X} + \sigma_{y}^{2}\mathbf{I} & \mathbf{K}_{X,*}\\
    \mathbf{K}_{X,*}^{T} & \mathbf{K}_{*,*}
    \end{pmatrix}\right)
\end{equation}
The predictive distribution for test points $\mathbf{X}_{*}$ is
\begin{align}
    p(\boldsymbol{f}_{*}|\mathcal{D}, \mathbf{X}_{*}) &= \mathcal{N}(\boldsymbol{f}_{*}|\boldsymbol{\mu}_{*|X}, \boldsymbol{\Sigma}_{*|X}) \\
    \boldsymbol{\mu}_{*|X} &= \boldsymbol{\mu}_{*} + \mathbf{K}_{X,*}^{\text{T}}(\mathbf{K}_{X,X} + \sigma_{y}^{2}\mathbf{I})^{-1}(\mathbf{y}-\boldsymbol{\mu_{X}}) \\
    \boldsymbol{\Sigma}_{*|X} &= \mathbf{K}_{*,*} - \mathbf{K}_{X,*}^{\text{T}}(\mathbf{K}_{X,X} + \sigma_{y}^{2}\mathbf{I})^{-1}\mathbf{K}_{X,*}
\end{align}
If the mean function is zero, the posterior mean can be written as
\begin{equation}
    \boldsymbol{\mu}_{*|X} = \mathbf{K}_{X,*}^{\text{T}}(\mathbf{K}_{X,X} + \sigma_{y}^{2}\mathbf{I})^{-1}\boldsymbol{y}
\end{equation}
For one test data point with $\boldsymbol{k}_{*}=[\mathcal{K}(\boldsymbol{x}_{*}, \boldsymbol{x}_{1}), \dots, \mathcal{K}(\boldsymbol{x}_{*}, \boldsymbol{x}_{N})]$, it becomes
\begin{equation} \label{eq:16}
    \mu_{*|X} = \boldsymbol{k}_{*}^{\text{T}}(\mathbf{K}_{X,X} + \sigma_{y}^{2}\mathbf{I})^{-1}\boldsymbol{y}
\end{equation}
The parameters of a kernel can be found by maximizing the log marginal likelihood given below
\begin{equation} \label{eq:17}
    \log p(\boldsymbol{y}|\boldsymbol{X},\boldsymbol{\theta}) = -\frac{1}{2}(\boldsymbol{y}-\boldsymbol{\mu}_{X})^{\text{T}}\mathbf{K}^{-1}_{\sigma}(\boldsymbol{y}-\boldsymbol{\mu}_{X}) - \frac{1}{2}\log |\mathbf{K}_{\sigma}| - \frac{N}{2}\log (2\pi)
\end{equation}
where $\mathbf{K}_{\sigma} = \mathbf{K}_{X,X} + \sigma_{y}^{2}\mathbf{I}$

\subsection{Kernel Ridge Regression}

Linear regression with an $\ell_{2}$ penalty on the weights is known as ridge regression. The objective function can be written as:
\begin{equation}
    \boldsymbol{w}^{*} = \underset{\boldsymbol{w}}{\text{argmin}} \sum_{n=1}^{N}(y_{n}-f(\boldsymbol{x}_{n};\boldsymbol{w}))^{2} + \lambda \|\boldsymbol{w}\|_{2}^{2}
\end{equation}
In the above equation, $f(x;w) = w^{T}x$. The vector containing optimal weights can be obtained as follows:
\begin{equation}
    \boldsymbol{w}^{*} = (\mathbf{X}^{T}\mathbf{X}+\lambda\mathbf{I})^{-1}\mathbf{X}^{T}\boldsymbol{y} = (\sum_{n=1}^{N}\boldsymbol{x}_{n}\boldsymbol{x}_{n}^{T}+\lambda\mathbf{I})^{-1}(\sum_{n=1}^{N}\boldsymbol{x}_{n}y_{n})
\end{equation}
To transform the above problem into kernel ridge regression (KRR), the function space view of the above problem should be considered where the positive definite kernel $\mathcal{K}$ defines the function space $\mathcal{F}$. The function space version of the above objective function can be defined as follows:
\begin{equation}
    f^{*} = \underset{f \in \mathcal{F}}{\text{argmin}} \sum_{n=1}^{N}(y_{n}-f(\boldsymbol{x}_{n}))^{2} + \lambda\|f\|^{2}
\end{equation}
It can be shown that  $f^{*}$ is the same as the posterior predictive mean of a GP\footnote{The GP and Kernel Ridge should ideally have the same performance owing to the same predictive model but they differ because the methods used to find the kernel hyperparameters are different; GP maximizes marginal likelihood using gradient descent and Kernel Ridge uses grid search coupled with cross-validation. The grid search approach tends to be slower in general.} as in Eq.~\ref{eq:16} \cite{pml2Book}. It is given as:
\begin{equation} \label{eq:21}
    f^{*}(\boldsymbol{x}_{*}) = \boldsymbol{k}_{*}^{T}(\mathbf{K}_{X,X}+\sigma_{y}^{2}\mathbf{I})^{-1}\boldsymbol{y}
\end{equation}

\subsection{Long Short-Term Memory}

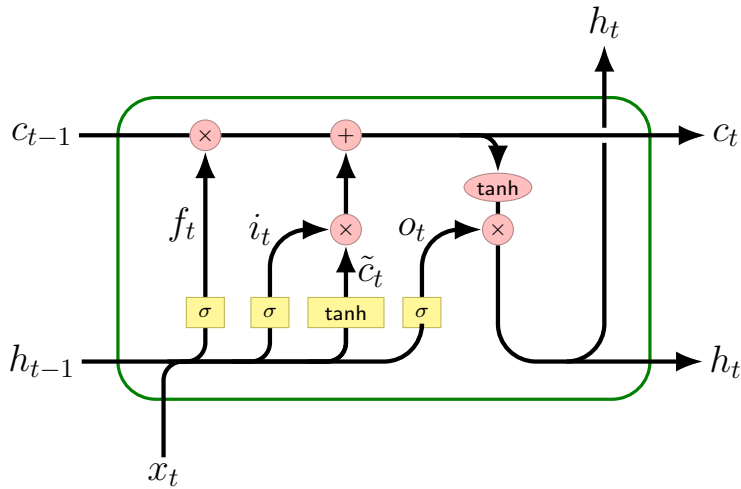
\begin{figure}[h]
  \centering
  \begin{tikzpicture}[
    % GLOBAL CFG
    font=\sf \scriptsize,
    >=LaTeX,
    % Styles
    cell/.style={% For the main box
        rectangle, 
        rounded corners=5mm, 
        draw=black!50!green,
        very thick,
        },
    operator/.style={%For operators like +  and  x
        circle,
        draw=black!30!pink,
        fill=pink,
        inner sep=1pt,
        minimum height =.2cm,
        },
    function/.style={%For functions
        ellipse,
        draw=black!30!pink,
        fill=pink,
        inner sep=1pt
        },
    ct/.style={% For external inputs and outputs
        %circle,
        %draw,
        line width = .75pt,
        inner sep=2pt,
        font=\large
        },
    gt/.style={% For internal inputs
        rectangle,
        draw=black!30!yellow,
        fill=white!50!yellow,
        minimum width=5mm,
        minimum height=4mm,
        inner sep=1pt
        },
    mylabel/.style={% something new that I have learned
        font=\scriptsize\sffamily
        },
    ArrowC1/.style={% Arrows with rounded corners
        rounded corners=.25cm,
        ultra thick,
        },
    ArrowC2/.style={% Arrows with big rounded corners
        rounded corners=.5cm,
        ultra thick,
        },
    ]

%Start drawing the thing...    
    % Draw the cell: 
    \node [cell, minimum height =4cm, minimum width=7cm] at (0,0){} ;

    % Draw inputs named ibox#
    \node [gt] (ibox1) at (-2.35,-0.85) {$\sigma$};
    \node [gt] (ibox2) at (-1.5,-0.85) {$\sigma$};
    \node [gt, minimum width=1cm] (ibox3) at (-0.5,-0.85) {tanh};
    \node [gt] (ibox4) at (0.5,-0.85) {$\sigma$};

   % Draw opérators   named mux# , add# and func#
    \node [operator] (mux1) at (-2.35,1.5) {$\times$};
    \node [operator] (add1) at (-0.5,1.5) {+};
    \node [operator] (mux2) at (-0.5,0.25) {$\times$};
    \node [operator] (mux3) at (1.5,0.25) {$\times$};
    \node [function] (func1) at (1.5,0.81) {tanh};

    % Draw External inputs? named as basis c,h,x
    %\node[ct, label={[mylabel]Cell}] (c) at (-4,1.5) {\empt{c}{t-1}};
    %\node[ct, label={[mylabel]Hidden}] (h) at (-4,-1.5) {\empt{h}{t-1}};
    %\node[ct, label={[mylabel]left:Input}] (x) at (-2.5,-3) {\empt{x}{t}};
    \node[ct] (c) at (-4.5,1.5) {\empt{c}{t-1}};
    \node[ct] (h) at (-4.5,-1.5) {\empt{h}{t-1}};
    \node[ct] (x) at (-2.9,-3) {\empt{x}{t}};

    % Draw External outputs? named as basis c2,h2,x2
    %\node[ct, label={[mylabel]Label1}] (c2) at (4,1.5) {\empt{c}{t}};
    %\node[ct, label={[mylabel]Label2}] (h2) at (4,-1.5) {\empt{h}{t}};
    %\node[ct, label={[mylabel]left:Label3}] (x2) at (2.5,3) {\empt{h}{t}};
    \node[ct] (c2) at (4.5,1.5) {\empt{c}{t}};
    \node[ct] (h2) at (4.5,-1.5) {\empt{h}{t}};
    \node[ct] (x2) at (2.9,3) {\empt{h}{t}};

% Start connecting all.
    %Intersections and displacements are used. 
    % Drawing arrows    
    \draw [ArrowC1] (c) -- (mux1) -- (add1);
    \draw [->, ArrowC2] (add1) -- (c2);

    % Inputs
    \draw [ArrowC2] (h) -| (ibox4);
    \draw [ArrowC1] (h -| ibox1)++(-0.5,0) -| (ibox1); 
    \draw [ArrowC1] (h -| ibox2)++(-0.5,0) -| (ibox2);
    \draw [ArrowC1] (h -| ibox3)++(-0.5,0) -| (ibox3);
    \draw [ArrowC1] (x) -- (x |- h)-| (ibox3);

    % Internal
    \draw [->, ArrowC2] (ibox1) -- node[ct,xshift=-0.12in] {$f_t$}(mux1);
    \draw [->, ArrowC2] (ibox2) |- node[ct,xshift=-0.05in] {$i_t$} (mux2);
    \draw [->, ArrowC2] (ibox3) -- node[ct,xshift=0.13in] {$\Tilde{c}_t$} (mux2);
    \draw [->, ArrowC2] (ibox4) |- node[ct,xshift=-0.05in] {$o_t$} (mux3);
    \draw [->, ArrowC2] (mux2) -- (add1);
    \draw [->, ArrowC1] (add1 -| func1)++(-0.5,0) -| (func1);
    \draw [-, ArrowC2] (func1) -- (mux3);

    %Outputs
    \draw [->, ArrowC2] (mux3) |- (h2);
    \draw (c2 -| x2) ++(0,-0.1) coordinate (i1);
    \draw [-, ArrowC2] (h2 -| x2)++(-0.5,0) -| (i1);
    \draw [->, ArrowC2] (i1)++(0,0.2) -- (x2);

\end{tikzpicture}
  \caption{An LSTM cell}
  \label{fig:lstm}
\end{figure}

For time-series forecasting, Long Short-Term Memory (LSTM) method introduced by Hochreiter et al. \cite{hochreiter1997long} has shown tremendous performance since its inception. The major component of it is a typical LSTM cell shown in Fig.~\ref{fig:lstm}.  
\\
There are three gates: input ($\boldsymbol{i}_t$), forget ($\boldsymbol{f}_t$), and output ($\boldsymbol{o}_t$) gates. Apart from this, there are hidden and cell states which helps the neural network to memorize the information in long time-series. The forget gate controls the amount of information fromm the cell state ($\boldsymbol{c}_{t-1}$) that will be carried over to the next cell state ($\boldsymbol{c}_t$), whereas the input gate controls the amount of information from the internal cell state ($\boldsymbol{\Tilde{c}}_{t}$) that will be carried over to the next state ($\boldsymbol{c}_t$). The next cell state is then determined by the summation of the previous cell state ($\boldsymbol{c}_{t-1}$) and ($\boldsymbol{\Tilde{c}}_{t}$) after they have been multiplied with their respective gates. The output gate determines the final value cell hidden features state ($\boldsymbol{h}_t$) based on the next cell state ($\boldsymbol{c}_t$). Also, the size of hidden features ($\boldsymbol{h}_t$) and cell state ($\boldsymbol{c}_t$) is very important as they not only determine the total number of parameters but the expressive power of LSTM model as well. The mathematical equations that govern an LSTM operation can be written as
\begin{align}
    \boldsymbol{i}_{t} &= \sigma(\mathbf{W}_{ii}\boldsymbol{x}_{t}+\boldsymbol{b}_{ii}+\mathbf{W}_{hi}\boldsymbol{h}_{t-1}+\boldsymbol{b}_{hi}) \\
    \boldsymbol{f}_{t} &= \sigma(\mathbf{W}_{if}\boldsymbol{x}_{t}+\boldsymbol{b}_{if}+\mathbf{W}_{hf}\boldsymbol{h}_{t-1}+\boldsymbol{b}_{hf}) \\
    \boldsymbol{\Tilde{c}}_{t} &= \text{tanh}(\mathbf{W}_{ig}\boldsymbol{x}_{t}+\boldsymbol{b}_{ig}+\mathbf{W}_{hg}\boldsymbol{h}_{t-1}+\boldsymbol{b}_{hg}) \\
    \boldsymbol{o}_{t} &= \sigma(\mathbf{W}_{io}\boldsymbol{x}_{t}+\boldsymbol{b}_{io}+\mathbf{W}_{ho}\boldsymbol{h}_{t-1}+\boldsymbol{b}_{ho}) \\
    \boldsymbol{c}_{t} &= \boldsymbol{f}_{t} \odot \boldsymbol{c}_{t-1} + \boldsymbol{i}_{t} \odot \boldsymbol{\Tilde{c}}_{t} \\
    \boldsymbol{h}_{t} &= \boldsymbol{o}_{t} \odot \text{tanh}(\boldsymbol{c}_{t})
\end{align}
In case of mulitlayer LSTM, such LSTM cells are stacked wherein the hidden state $\boldsymbol{h}_{t}^{l-1}$ of the previous layer becomes the input $\boldsymbol{x}_{t}^{l}$ of the $l^{th}$ layer for $l \geq 2$. We should note that the parameters of an LSTM cell in the same layer are shared across timesteps; this makes it invariant to the size of input sequence. 

In this section, a few classical car-following and data-driven regression models have been described briefly---a brief overview of the characteristics of those models is given here. These models can be broadly divided into two categories: parametric and non-parametric. All classical models and LSTM are parametric models, whereas GP and Kernel Ridge are non-parametric models. 
\begin{itemize}
    \item In the case of parametric models, parameters in some models like IDM, Gipps, and FVDM have a certain physical meaning, whereas parameters in LSTM have no intuitive physical meaning. For instance, IDM has the following parameters: desired velocity ($V_{max,n}$), minimum gap ($s_{0,n}$), time headway ($T_{n}$), acceleration ($a$), and deceleration ($b_n$); and the following inputs: $v_n$, $s_n$, and $\Delta v_{n}$. Other models like Gipps and FVDM (and its two variants) have their own respective parameters and inputs.
    \item In the case of non-parametric models such as GP and Kernel Ridge Regression, there are no parameters, and they need access to the dataset for inference; in contrast, parametric models don't assume access to the dataset once they are trained and need model parameters only for predictions.
\end{itemize} 
All these models do the same job of modeling the relationship between inputs and output. In the case of data-driven models (GP, Kernel Ridge, and LSTM), the inputs are always $v_n$, $s_n$, and $v_{n-1}$, where $n-1$ indicates the vehicle directly in front of the vehicle $n$; and the output could be one of the following: $a_n$, $v_n$, and $s_n$, depending on the design variables (Appendix \ref{sec:rmse}). On the other hand, all classical models described in this section output $a_n$ and take the same inputs -- following vehicle speed, $v_n(t)$, intervehicle spacing, $s_n(t)$, and relative velocity, $\Delta v_n(t)$ -- except Gipps, which needs an additional input ($v_{n-1}$).

\section{Results}\label{sec:resultss}

As discussed in section \ref{sec:related}, the choice of a target variable impacts the predictive performance of the model---both symbolic (e.g. IDM) and black-box (e.g. LSTM). This section will show that the effects of target variables vary in nature depending on the model under consideration; these varying effects result in different suggestions regarding target variable selection for different models.

Three datasets are used in the paper: ASTAZERO \cite{makridis2021openacc}, JIANG \cite{jiang2015some}, and NAPOLI \cite{punzo2005analysis}. ASTAZERO was collected via automated vehicles driving in a platoon in Sweden. JIANG is a collection of trajectories from human-driven vehicles, and the experiments were carried out in Hefei City. NAPOLI was collected from human-driven vehicles in the city of Naples.

First, we discuss the performance of all models corresponding to different target variables selection. Second, results from analysis of variance (ANOVA) is presented to elaborate on the interplay of different types of variables -- target, models, and datasets -- and how the different choices of these variables  can impact the RMSE of predictive variables. 

\subsection{CF Models Performance}

\subsubsection{Implementation Details}

The symbolic forms of the four CF models under consideration are given in section 3; the parameters of each CF model are indicated in bold. The values of these parameters vary depending on the dataset and can be found by optimizing a CF model using optimization algorithms \cite{brockfeld2003toward}; a comprehensive overview of the optimization algorithms used for CF model calibration is given in \cite{fard2019copula}. Only a few optimization algorithms find optimal parameters for a CF model, namely gradient-based NLP solvers and genetic algorithms (GA), whereas other algorithms (Trust Region, Interior Point, and Sequential Quadratic Programming) could not find optimal parameters \cite{punzo2012can}.

\begin{figure}[h]
  \centering
    \includegraphics[scale=0.56]{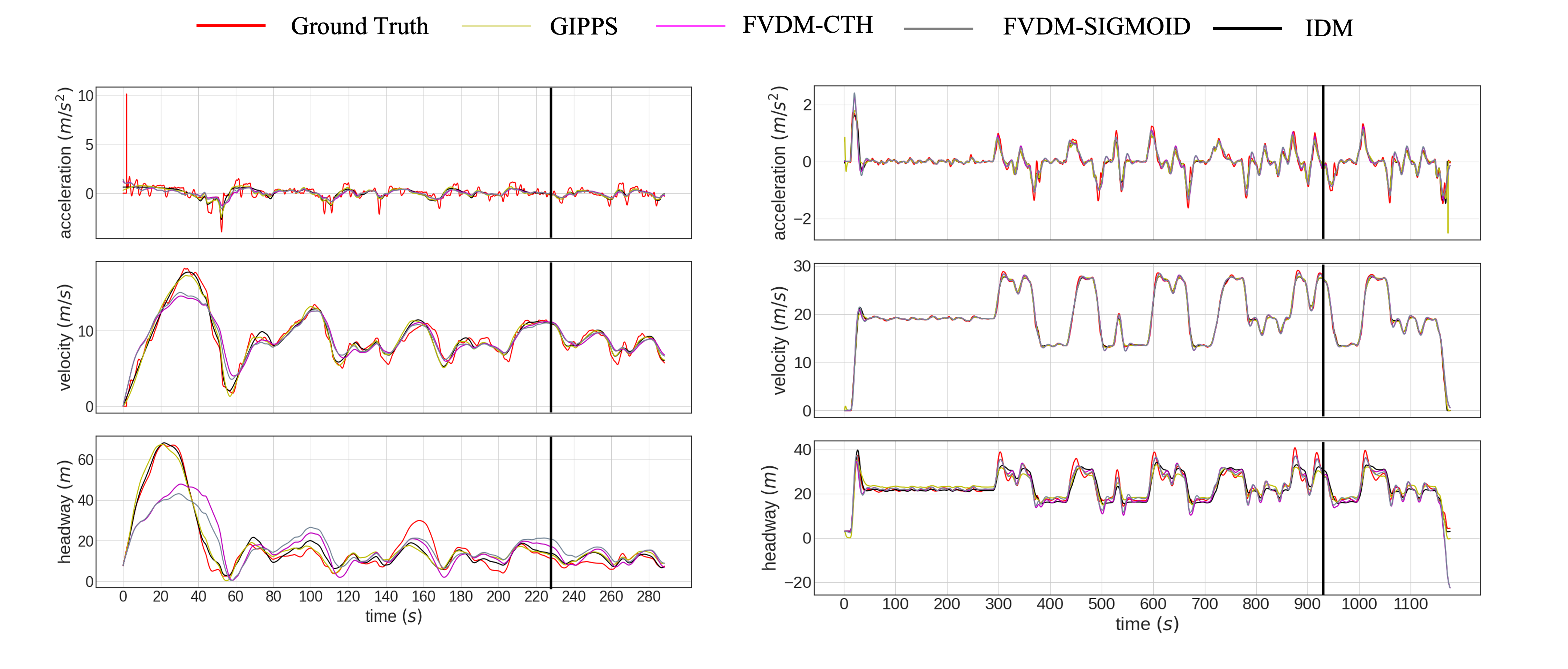}
  \caption{The performance of four classical CF models with their respective learned parameters on (a) ASTAZERO ($s$) (b) JIANG ($s$). \protect\footnotemark \hspace{0.01cm} The black line on each plot indicates the 80/20\% split between training and test data respectively.}
  \label{fig:classical(s)}
\end{figure}

Therefore, for this study, GAs implemented in $\mathtt{Matlab}$ were used to calibrate the CF models. The code given in \cite{punzo2021calibration} was used with a few modifications. The primary modification relates to the notion of training and testing data. In general, the data-driven algorithms are trained on the training data to learn the underlying dynamics and then tested on the testing (unseen) data. The same methodology was used in the present work for GP, KRR, and LSTM. The same approach towards classical CF models is thus necessary if we use classical CF model results as a benchmark against data-driven models. To that end, the time-series were divided into two parts: 80\% for training and 20\% for testing. The training data was used to find the optimal parameters of a classical CF model, and then the calibrated parameters were used for testing against the testing data. The performance of ASTAZERO and JIANG with target variable $s$ is shown in Fig.~\ref{fig:classical(s)}.\\
\footnotetext{The target variables are given in parenthesis after dataset name.}

\subsubsection{Performance Analysis}
As shown in Fig.~\ref{fig:classical(s)}, the parameters optimized on target variable $s$ give the best performance in terms of RMSEs across three variables ($a$, $v$, and $s$), whereas the performance with target variables $v$ and $a$ is relatively poor. The performance of classical models with target variables $v$ and $a$ is shown in Appendix \ref{sec:classical(a & v)}. The RMSEs are given in Appendix \ref{sec:rmse}. Different target variables (with RMSE-based objective function) were used by different authors: RMSE ($s$) \cite{xu2020statistical,hammit2018evaluation,park2019development}, RMSE ($v$) \cite{gunter2019model,li2020car,ye2018prediction}, and RMSE ($a$) \cite{pei2016empirical,xu2018aware,hao2016fuzzy}. Most papers reported superior performance of CF models when optimized on $v$ or $s$; $s$ has been reported the most often as a best target variable \cite{punzo2021calibration}. One recent paper \cite{punzo2016speed} focused on finding the reasoning behind the superior performance of $s$ over $v$; equations~\ref{eq:28} and~\ref{eq:29} explain that---the derivation can be found in the original paper. When a model is optimized on $s$, it is optimal in headway ($s$) and suboptimal in velocity ($v$) (Eq.~\ref{eq:28}). On the other hand, a model optimized on velocity ($v$) is optimal in velocity ($v$) but indeterminate in headway ($s$); that is, the different errors in spacing can correspond to the same error in speed (Eq.~\ref{eq:29}). It is worth noting that the reasoning based on Eqs.~\ref{eq:28} \&~\ref{eq:29} is true for sum of square (SSE) based objective functions, like RMSE, etc. However, this may not work for other objective functions (e.g. negative log marginal likelihood).
 \begin{align} 
    \text{min}\{SSE^{V}\} &= \text{min}\biggl\{\sum^{N}_{k=1}(\epsilon^{V}_{k})^{2}\biggl\} \label{eq:28}\\
    \text{min}\{SSE^{S}\} = \text{min}\biggl\{\sum^{N}_{k=1}(\epsilon^{S}_{k})^{2}\biggl\} &= \text{min}\biggl\{\sum^{N}_{k=1}(\epsilon^{V}_{k})^{2} + \left(k^{*}\left(\varepsilon^{V}_{k}\right)^{2}\right)[N] + 2\sum^{N-1}_{k=1}\varepsilon^{V}_{k}.\sum^{N}_{i=k+1}\left(N-i+1\right).\varepsilon^{V}_{i}\biggl\} \label{eq:29}
\end{align}

\subsection{Gaussian Process \& Kernel Ridge Regression Performance}

\subsubsection{Brief description of Model}
GP and Kernel Ridge have the same predictive mean as shown in section 3.1 \& section 3.2 (Eqs.~\ref{eq:16} and~\ref{eq:21}), with GP having an added benefit of predictive uncertainty. We note that the uses of Gaussian Processes far exceed that of Kernel Ridge Regression in the literature and hypothesize several reasons. The first is that Gaussian Processes provide the same prediction as Kernel Ridge Regression plus an additional term, the covariance prediction. A second reason could be the lack of a proper method of finding kernel parameters for KRR, other than grid search coupled with cross-validation; it can be seen that this causes major performance degradation compared to GP (Appendix B). 

The results of the present paper will be compared with the two other studies \footnote{Since different papers use different datasets, a side-by-side performance comparison is not possible. Instead, a few general comments regarding the target variable and kernel selection will be made.}: First, a study by Soldevila et al. \cite{soldevila2021car} wherein GP was used to model car-following dynamics and optimal velocity model (OVM) model was used as a mean instead of a constant mean in GP. Second, a study by Han et al. \cite{wang2021personalized} that used GP to model car-following dynamics based on simulated (IDM-based) and naturalistic (collected using Logitech racing wheel and Unity engine) data. While comparing, some of the pitfalls of the improper kernel and target variable selection will be highlighted. 

\subsubsection{Details about implementation}\label{sec:DetImpGP}
 Different kernels were tested (both with GP \& Kernel Ridge) instead of simply relying on the most ubiquitous radial-basis function (RBF); these include rational quadratic, exponential, Matérn, and multi-layer perceptron (MLP) aka arc sine. The mathematical forms of these kernels are given in Appendix B. The kernel with the best performance (in terms of RMSE) across all three variables ($a$, $v$, and $s$) was chosen as a baseline. Interestingly, the kernel with the best performance in most cases were Matern, exponential, and MLP rather than RBF. It is unclear why certain kernels perform well on specific trajectories and target variables and this warrants further research. Regarding implementation, $\mathtt{GPy}$ \cite{hensman2012gpy} and $\mathtt{sklearn}$ \cite{scikit-learn} implementations of GP and Kernel Ridge were used, respectively. The code is available online \footnote{\url{https://github.com/coordinated-systems-lab/CarFollowModels}}. 

 \subsubsection{Performance Analysis}\label{sec:PerfAnalGP}
\begin{figure}[h]
    \begin{minipage}[h]{0.47\linewidth}
        \begin{center}
        \includegraphics[width=1.13\linewidth]{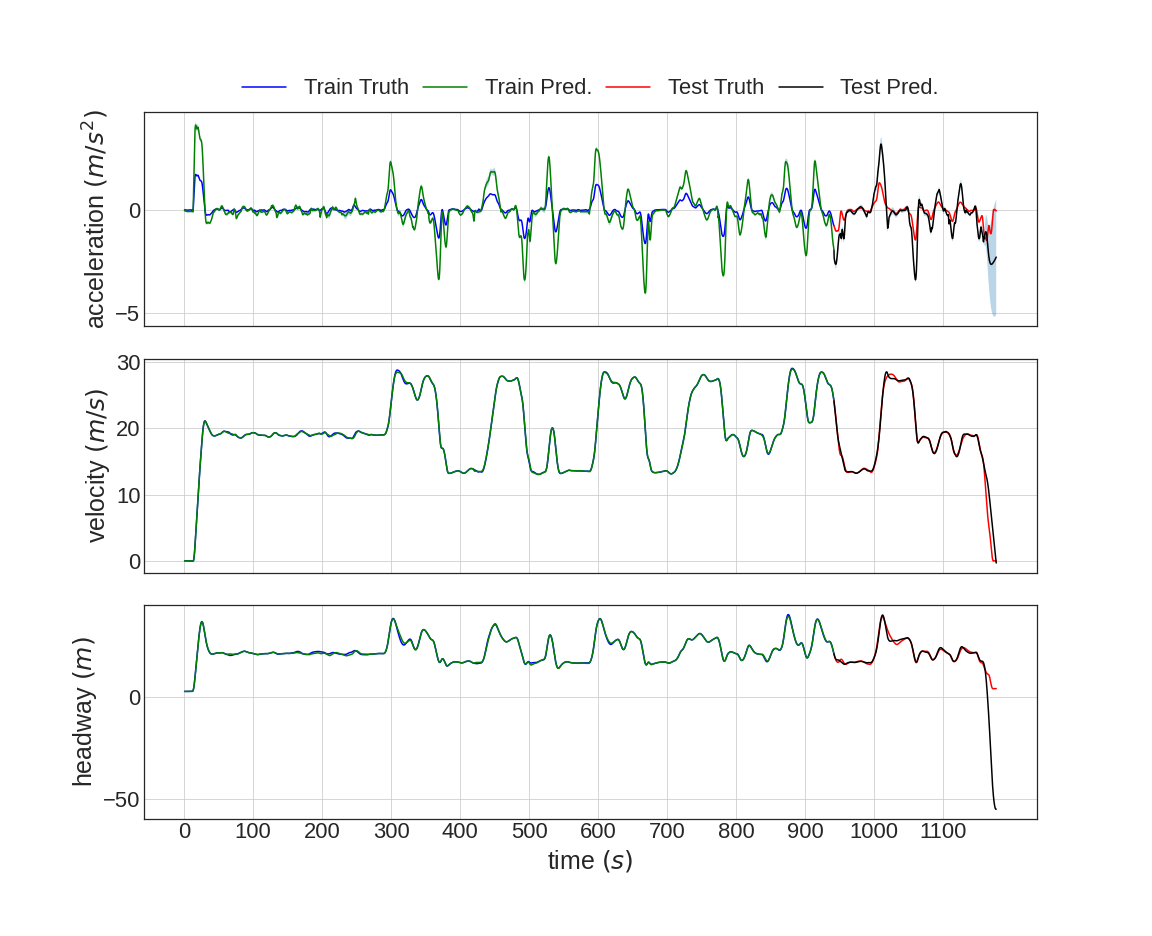}
        \captionsetup{justification=centering, margin=4.4cm, labelformat=empty}
        \vspace*{-11mm}
        \subcaption{\label{fig:GPAsta(a)}}
        \end{center} 
    \end{minipage}
    \begin{minipage}[h]{0.47\linewidth}
        \begin{center}
        \includegraphics[width=1.13\linewidth]{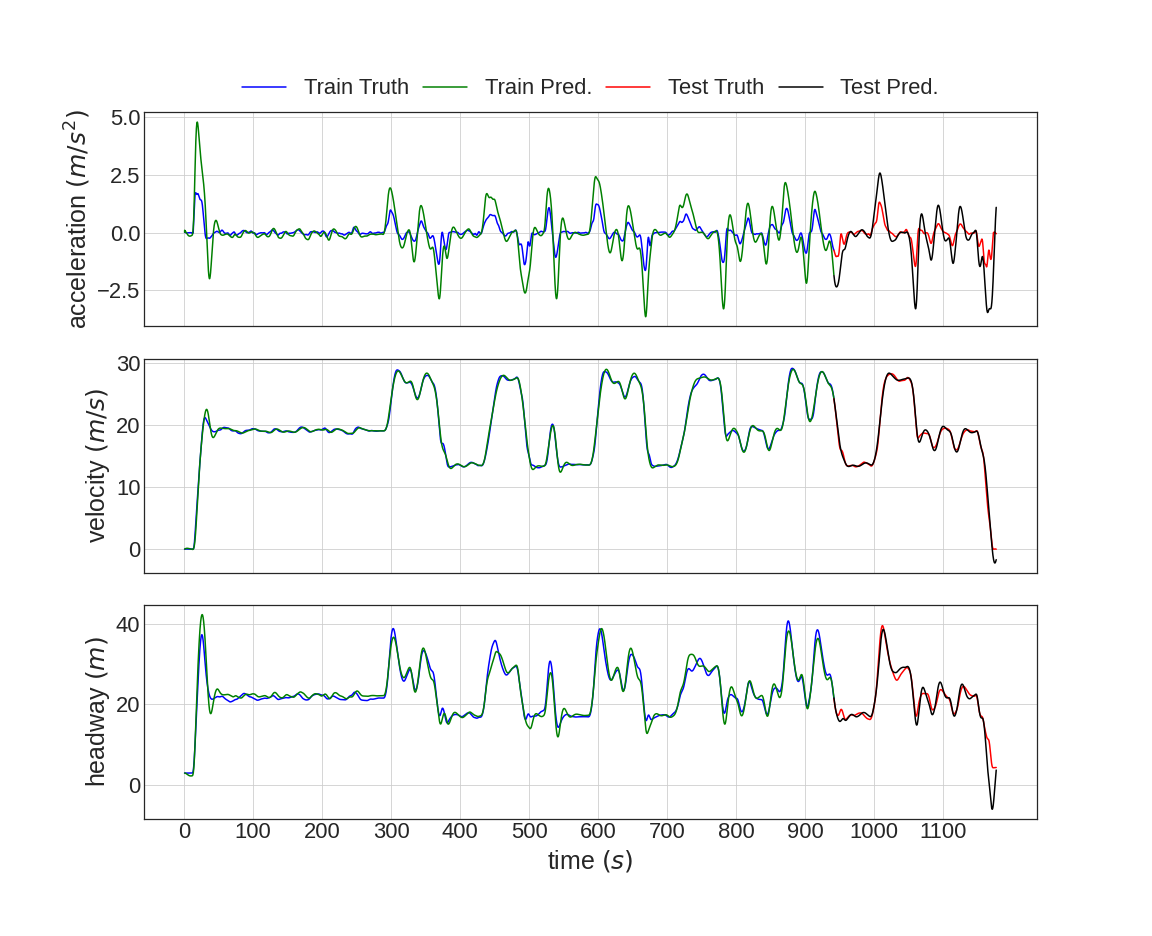}
        \captionsetup{justification=centering, margin=4.4cm, labelformat=empty}
        \vspace*{-11mm}
        \subcaption{\label{fig:KRRAsta(a)}}
        \end{center}
    \end{minipage}
    \vfill
    \vspace*{-1.2mm}
    \begin{minipage}[h]{0.47\linewidth}
        \begin{center}
        \includegraphics[width=1.13\linewidth]{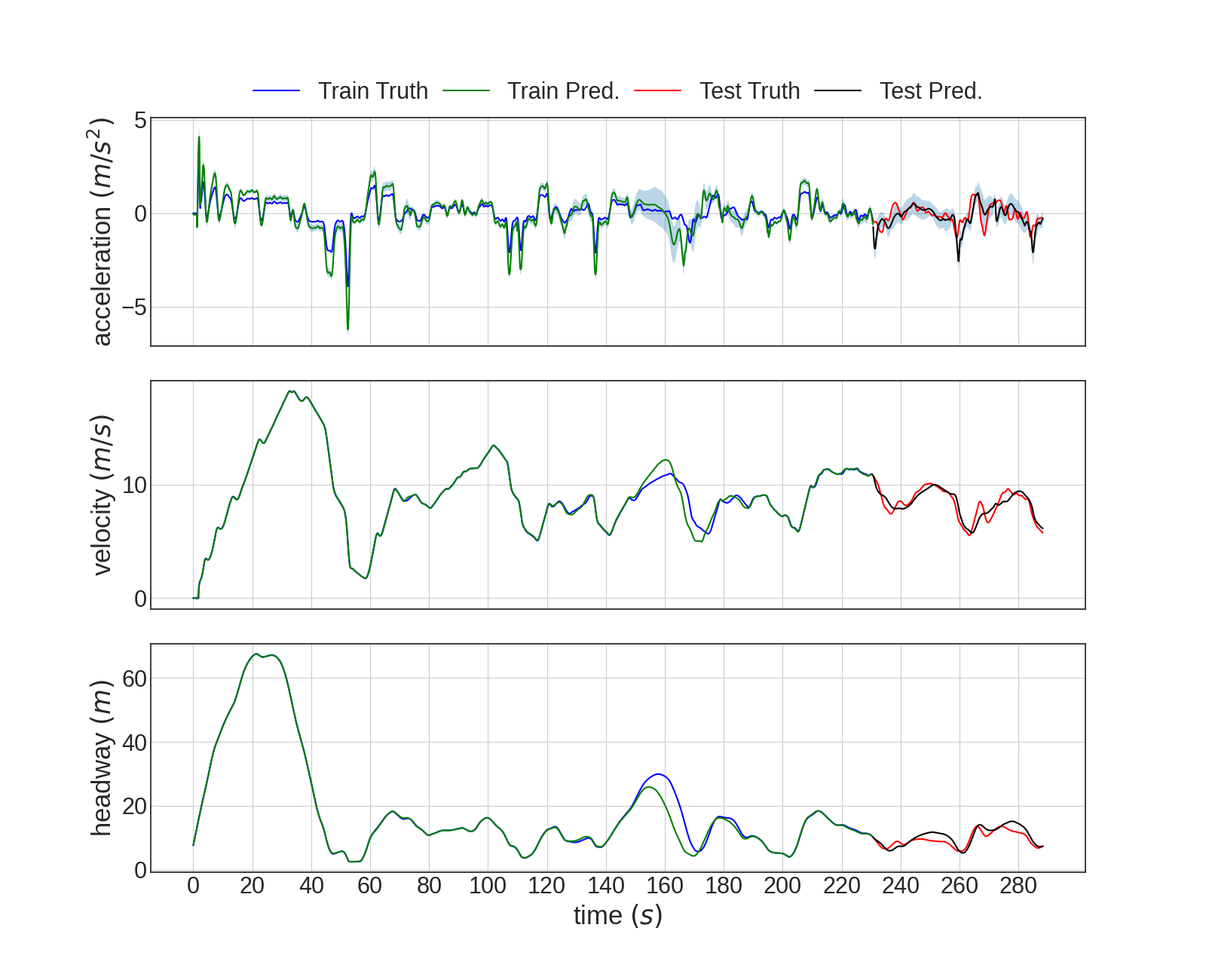}
        \captionsetup{justification=centering, margin=4.4cm, labelformat=empty}
        \vspace*{-11mm}
        \subcaption{\label{fig:GPJiang(a)}}
        \end{center}
    \end{minipage}
    \begin{minipage}[h]{0.47\linewidth}
        \begin{center}
        \includegraphics[width=1.13\linewidth]{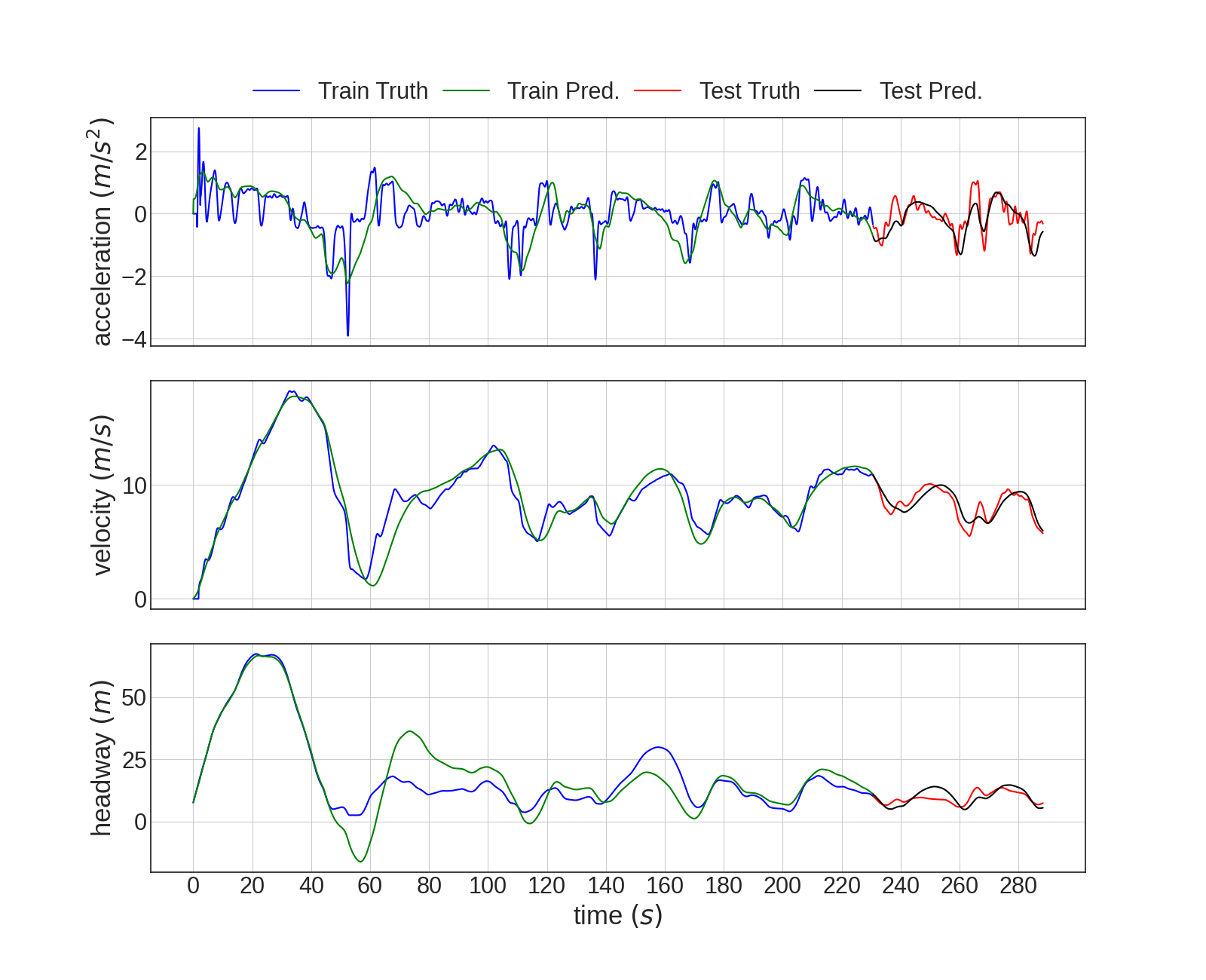} 
        \captionsetup{justification=centering, margin=4.4cm,labelformat=empty}
        \vspace*{-11mm}
        \subcaption{\label{fig:KRRJiang(a)}}
        \end{center}
    \end{minipage}
    \vspace*{-3mm}
    \caption{The performance of GP \& KRR on the ASTAZERO and JIANG datasets with $a$ as the target variable (a) GP on ASTAZERO ($a$) (b) KRR on ASTAZERO ($a$) (c) GP on JIANG ($a$) (d) KRR on JIANG ($a$)}
    \label{fig:GPandKRR(a)}
\end{figure}

 The GP and Kernel Ridge results on ASTAZERO and JIANG with target variable $a$ are shown in Fig.~\ref{fig:GPandKRR(a)}. The results of best performing target variable (i.e. $a$) on two datasets are given in this section, while the performances of other target variables are given in appendix A.2. The best-performing target variable in the case of GP and Kernel Ridge is $a$. Fig.~\ref{fig:GPandKRR(a)} is in direct contrast with the results shown in Fig.~\ref{fig:classical(s)} for classical CF models where the best performing target variable is $s$. It could be due to two main reasons: first is the objective function of two approaches (GP and GA) and is based on the Eq.~\ref{eq:28} \&~\ref{eq:29}, which explains the better performance of $s$ while using the SSE-based objective functions; second reason could be the nature of the problem itself---GP and Kernel Ridge work in function spaces (Reproducing Kernel Hilbert Spaces) and GA works in parameter space. We should note that Kernel Ridge also uses the SSE-based objective function, but it works in function space rather than parameter space; hence it gives different results than other algorithms that use the SSE-based objective function. Due to the above reasons, GA performs better with $s$ as the target variable, whereas GP and Kernel Ridge perform better with $a$ as a target variable. In the case of kernel-based methods (GP and Kernel Ridge), $a$ and $v$ are the preferred target variables, while $s$ should be avoided. On the other hand, $v$ and $s$ are the preferred target variables for the SSE-based objective functions while $a$ should be avoided \cite{punzo2012can,punzo2021calibration}. It is worth noting that the results for GP, Kernel Ridge, and LSTM were generated in \textbf{free simulation mode}; that is, the input features at the next time-step were updated based on the predicted output from the last time-step. This is not a trivial task as the errors in prediction accumulate over time, and the resulting trajectory drifts away from ground truth---some performance improvements can be made as discussed in \cite{venkatraman2015improving, ross2011reduction}. This was done to make the comparison of these models fair with the classical CF models.\\
 
Emphasis should be put on trying different kernels while using GPs, as this choice impacts the multi-step predictive performance of the learned model across all variables ($a$, $v$, and $s$). The multi-step prediction is a harder task than one-step prediction because the predictions of the model are fed back as inputs to get the prediction at next time-step and predictions drift away from the ground truth over time; one such example is shown in Fig.~\ref{fig:classicalASTA(a)} where the model was optimized on $a$ but the small inaccuracies in prediction of $a$ accumulated over time during integration resulting in a large error in $s$. All kernels resulted in good one-step prediction, but only a few of them resulted in good multi-step predictions. Those kernels were Exponential, Matern, and MLP, whereas RBF worked only in a very few cases. Soldevila et al. \cite{soldevila2021car} used RBF kernel and reported several collisions during simulation of the learned model, whereas there are many cases in the present study with positive headway (i.e. zero collision) that use a different kernel (i.e. other than RBF). Another study \cite{wang2021personalized} reported a very low mean squared error of 0.0553 on acceleration---the target variable---and relatively higher MSE of 81.6 on $s$ (Table I in \cite{wang2021personalized}); we believe that this is owing to improper selection of the kernel. The same paper reported the results along the same lines for IDM (i.e. lower MSE on $a$ but higher on $s$) and this is owing to improper selection of target variable. To support this argument, ANOVA results are presented in section 4.4 wherein the Fig.~\ref{fig:anovaRMSE}b \& ~\ref{fig:anovaRMSE}d show that there is no interaction between Data \& Model and Data \& Target and selection of target variables depends only on the model (e.g., GP, Kernel Ridge, and LSTM) under consideration irrespective of the dataset. \\

\subsection{LSTM Performance}

\subsubsection{Brief description of model}
LSTM networks with varying numbers of hidden layers and neurons in each layer were built to optimize performance on the given data sets---one dense layer was used at the end for all architectures to convert hidden features of the last LSTM layer to final predictions. Three-layer stacked LSTM (25 neurons in each layer) with one dense layer at the end was finally chosen for training across all data sets. The input feature variables were: the follower's velocity ($v$), the leader's velocity ($v_{l}$), and headway ($s$); the predicted variable was different in each case depending on whatever target variable was used for optimization. For instance, if $a$ was used as a target variable to optimize the objective function, the predicted variable was the same as well. Since the predicted variable is a real number, a sequence-to-real (Seq-to-Real) LSTM implemented in PyTorch was used. Since it's a Seq-to-Real LSTM, the past history consisting a few timesteps (5 in this case) of input variables should be used as features; by contrast, only the features at the last time-step were used as input to GP and KRR. The history of past time-steps to be used as a feature was selected as 5 after different trials. The objective function used during training was MSE. 

\subsubsection{Details about implementation}
\begin{figure}[h]
  \centering
  \begin{minipage}[b]{0.47\linewidth}
    \includegraphics[width=1.12\linewidth]{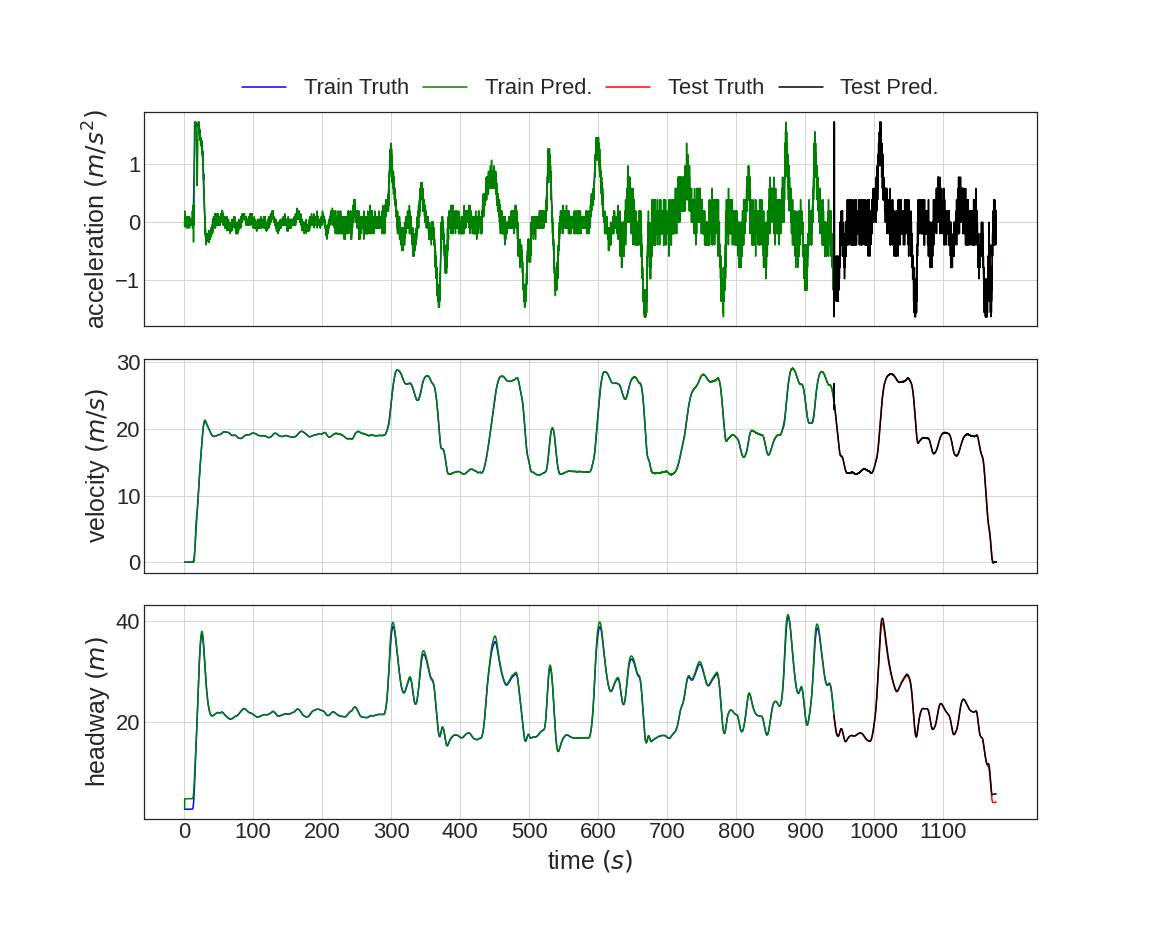}
    \captionsetup{justification=centering, margin=4.4cm, labelformat=empty}
    \vspace*{-10mm}
    \subcaption{\label{fig:LSTMAsta(s)}}
  \end{minipage}
  \begin{minipage}[b]{0.47\linewidth}
    \includegraphics[width=1.12\linewidth]{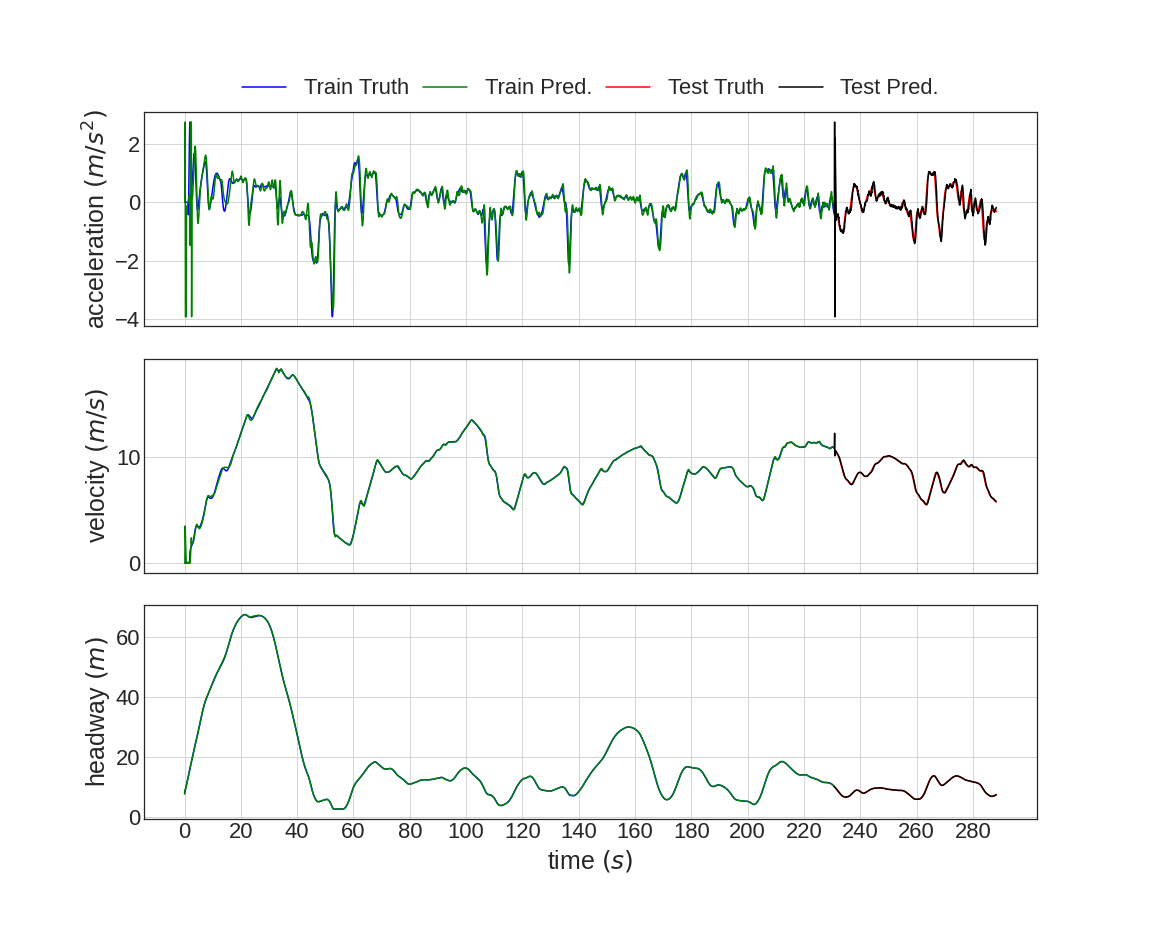}
    \captionsetup{justification=centering, margin=4.4cm,labelformat=empty}
    \vspace*{-10mm}
    \subcaption{\label{fig:LSTMJiang(s)}}
  \end{minipage}
  \vspace*{-3mm}
  \caption{The performance of LSTM on (a) ASTAZERO ($s$) (b) JIANG ($s$)}
  \label{fig:LSTM(s)}
\end{figure}

The performance of LSTM on ASTAZERO and JIANG with $s$ as a target variable is shown in Fig.~\ref{fig:LSTM(s)}. To see the performance of other target variables on both datasets, please refer to appendix A.3. Just like in previous sections, the results of the best-performing target variable $s$ are shown in here. The best-performing target variable in the case of classical CF models and LSTM is the same, and it is because both models used the same objective function (SSE) and work in parameter space. As shown in Eqs.~\ref{eq:28} \&~\ref{eq:29} \cite{punzo2016speed}, the optimization of the SSE-based objective function on $s$ results in better performance across all three variables. Since both GA and LSTM used SSE-based objective functions, their performance with different target variables is the same. As shown in Fig.~\ref{fig:classicalASTA(a)} and~\ref{fig:LSTMAsta(a)}, the performance of the models (classical CF and LSTM) optimized on target variable $a$ is good on $a$ itself but it performed poorly on both $v$ and $s$ because the error accumulates during integration when $a$ is integrated to get $v$ and $s$; the same is true when model is optimized on $v$, i.e., it performs very well on $v$, but it performs poorly when it is integrated to obtain $s$ (Fig.~\ref{fig:classicalASTA(v)} and~\ref{fig:LSTMAsta(v)}). Hence, the results shown in Figs.~\ref{fig:classical(s)},~\ref{fig:LSTM(s)},~\ref{fig:classical(v)and(a)}, and~\ref{fig:LSTM(v)and(a)} along with the same recommendations on target variable selection shows that the classical CF models and LSTM exhibited similar behavior owing to same SSE-based objective functions. 

\subsubsection{Performance Analysis}\label{sec:PerfAnalLSTM}
Two papers used approaches similar to the one discussed in this section, and our results are compared to theirs to justify the better performance of $s$ as both target and predicted variable: First, Zhou et al. \cite{zhou2017recurrent} used recurrent neural networks (RNN) with input features ($v$, $s$, $\Delta v$ \footnote{The difference of leader and follower velocity}) and predicted variable as $a$; the target variable was $s$. Second, Huang et al. \cite{huang2018car} uses the same input features with predicted variable $v$; the target variables were $v$ and $s$. The reported MSEs in both papers on variables vary depending on the situation, but in most cases, the reported MSEs are high compared to our study. A side-by-side comparison is not possible because of different data sets, but the results can still be compared based on reported MSEs. The MSEs on follower position reported by Zhou et al. \cite{zhou2017recurrent} can be as high as 115 and as low as 1.82 (Fig. 16, 17, 18 in \cite{zhou2017recurrent}), whereas the MSEs reported by Huang et al. \cite{huang2018car} range within 0.62 and 0.99 for $v$ and 1.89 and 212.88 for $s$ (Table 6 in \cite{huang2018car}). On the other hand, RMSEs from our current study based on LSTM reported in Appendix \ref{sec:rmse} are very low for target variable $s$ (range between 0.079 and 0.41). Despite the differences in data sets, we believe that the results with $s$ as both the target and predicted variable can be improved compared to the choices made by Zhou and Huang. To support this argument, ANOVA results are presented in section \ref{sec:anova} wherein the Fig.~\ref{fig:anovaRMSE}b \&~\ref{fig:anovaRMSE}d show that there is no interaction between Data \& Model and Data \& Target and selection of target variables depends only on the model (e.g., GP, Kernel Ridge, and LSTM) under consideration irrespective of the dataset.

The previous subsections compare the performance of different models based on different target variables. Another evaluation criterion is fundamental diagrams, which describe the relationship between traffic flow, density, and speed on a roadway. Since CF models directly impact the aforementioned factors, fundamental diagrams have been used by many as a performance evaluation criterion. However, fundamental diagrams are out of scope for the following reasons. First, GPs, LSTM, etc. have been studied from this perspective before \cite{cheng2022bayesian,ma2020sequence}. Second, one component lacking from most studies is a comprehensive evaluation of proper target variables for black-box models, which is the goal of this paper.

\subsection{Robustness of recommended target variables}\label{sec:anova}
In the present work, there were at least three sources of variation in the design of experiments (shown in Appendix \ref{sec:rmse}), namely, Dataset, Model, and Target variable, i.e. {\em independent} variables. The RMSEs of predicted variables (i.e., $a$, $v$, and $s$), or {\em dependent} variables, varied substantially, corresponding to different combinations of independent variables. ANOVA is used to elaborate on the effect of change in independent variables on dependent variables.

Another variable that might affect results is the optimization algorithm for a model. There are different optimization algorithms, and the best option depends on the problem under consideration. There were four different models in this study: classical CF models, Gaussian Process, Kernel Ridge Regression, and LSTM. The classical CF models were optimized using GA, which has been the most effective at optimizing CF model parameters \cite{fard2019copula}. Gaussian Processes are non-parametric models with hyperparameters depending on the kernel. The hyperparameters were tuned using an L-BFGS-B optimizer via maximum likelihood estimation (MLE). Kernel ridge regression involves finding coefficients of kernelized input features. It is a linear system that can be efficiently solved using Cholesky decomposition. We used an implementation of kernel ridge regression from $\mathtt{sklearn}$ \cite{scikit-learn}. Finally, LSTM parameters were optimized using a stochastic gradient descent optimizer available in $\mathtt{PyTorch}$ \cite{paszke2019pytorch}. In some cases, it would have been possible to implement different optimization methods on the models, but in others, it is either not practical or theoretically impossible (for example, Cholesky decomposition is simply not an option for LSTMs, nor is GA). We, therefore, used best practices from the literature for optimization of each class of models and do not include this variable in the ANOVA.
%Each of these models is optimized with different methods and optimizers with varying performances, but this variability is inevitable as it is challenging (and in some cases theoretically infeasible) to select one optimizer that works equally well for all models. Therefore, it was not considered a source of variation in the ANOVA. 

\begin{figure}[htb!]
  \centering
  \includegraphics[width=\linewidth,trim={4.5cm 0 4.5cm 0},clip]{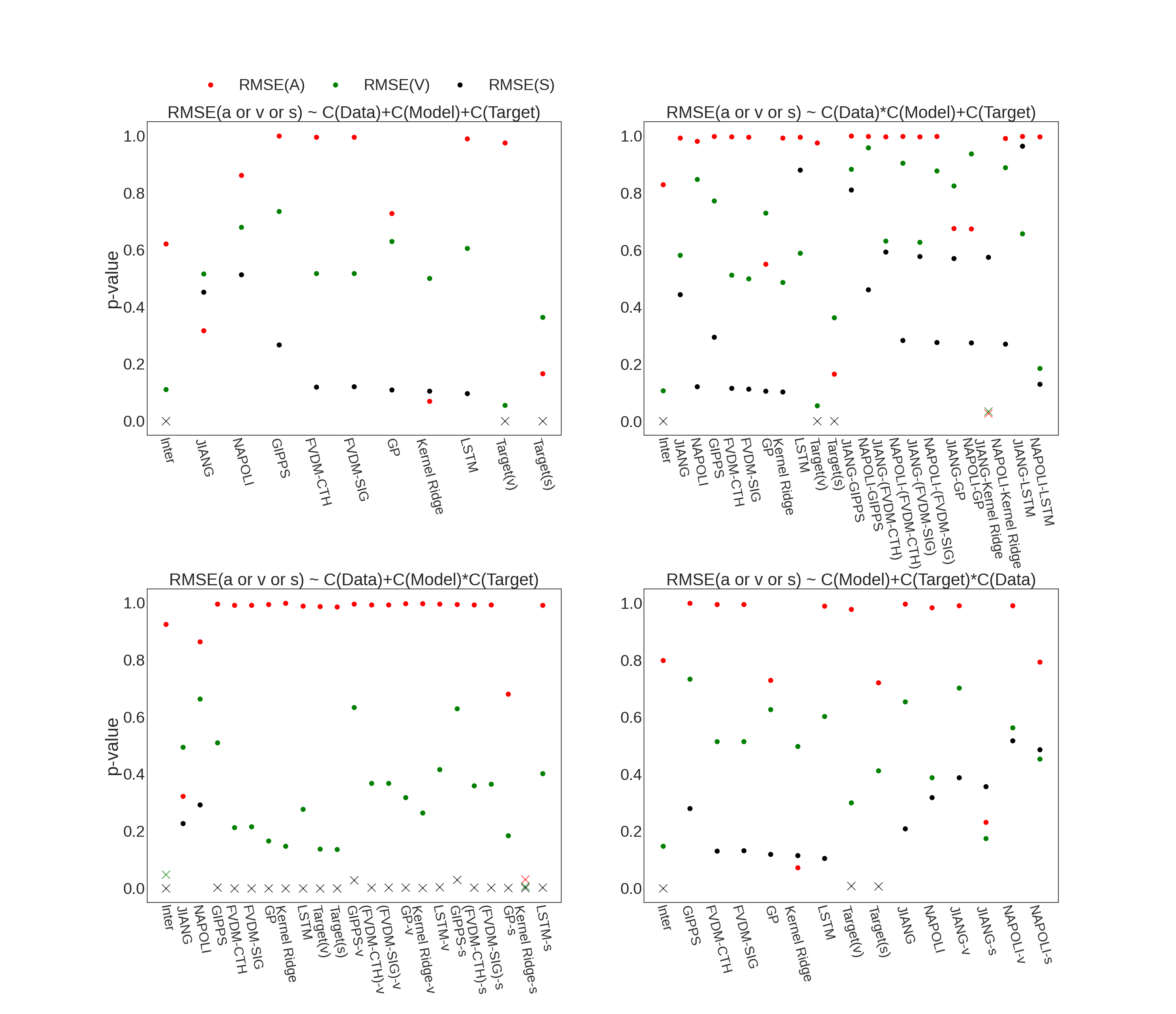}
  \caption{The plots with p-values showing (a) the effect of independent variables on dependent variables (b) the interaction between Data and Model (c) the interaction between Model and Target (d) the interaction between Data and Target}
  \label{fig:anovaRMSE}
\end{figure}

First, the effect of each independent variable is seen on three dependent variables, and p-values are plotted in Fig.~\ref{fig:anovaRMSE}a; the variables with p-values less than 0.05 are believed to have a considerable effect on the dependent variables and shown with the cross symbols as such. It can be seen that the \textit{Target} impacts the predicted variables RMSEs the most, which implies that the selection of target variables is crucial. Second, the interaction of independent variables was considered; the interactions of Data \& Model, Model \& Target, and Data \& Target are shown in Fig.~\ref{fig:anovaRMSE}b,~\ref{fig:anovaRMSE}c, and~\ref{fig:anovaRMSE}d, respectively. The three interactions are explained below briefly.
\begin{itemize}
    \item In Fig.~\ref{fig:anovaRMSE}b, it can be seen that the Data \& Model are not interacting with each other with one exception (i.e., JIANG-Kernel Ridge); the reason lies in the fact that the grid search method sometimes failed to find optimal kernel parameters for Kernel Ridge. The fact that the GP performed well in the same situation endorses this argument.
    \item In Fig.~\ref{fig:anovaRMSE}c, a lot of p-values (less than 0.05) can be seen represented by the crosses, which shows that the Model \& Target interacts in many cases. This corroborates the results presented in the last sections, wherein the effect of the target variable was shown on the predicted variable for different models (e.g., GP, KRR, LSTM). For instance, the target variable $s$ gives the worst performance for GP \& KRR (Fig.~\ref{fig:GPandKRR(s)}), whereas it gives the best performance for LSTM (Fig.~\ref{fig:LSTM(s)}). 
    \item In Fig.~\ref{fig:anovaRMSE}d, the crosses can be seen only against the target variables (i,e., Target(v) and Target(s)); this shows that the Data \& Target do not interact at all which means that the recommendations regarding target variable selection depend on the model irrespective of the dataset under consideration. \textit{Similar reasons and rationale legitimate the performance comparisons made in sections \ref{sec:PerfAnalGP} \&  \ref{sec:PerfAnalLSTM} -- wherein the results from other papers were compared to the current study and recommendations regarding target variable selection were made even though different papers used different datasets}.    
\end{itemize}

The results of different data-driven models (GP, Kernel-Ridge, and LSTM) on different datasets have been presented, and performance comparisons were made with a particular focus on selecting optimal target variables. The nature of the model can also impact performance. It was observed that the LSTM performs better than GP in terms of RMSE. The RMSEs of all datasets with all target variables are shown on $\log$ scale in \ref{sec:logRMSE} with original values in \ref{sec:rmse}. The LSTM has the lowest RMSE across $a$, $v$, and $s$ with the target variable $s$ for all datasets. The primary reason for this performance difference is that LSTM is a pattern discovery algorithm with its expressive parametric representation. On the other hand, GP acts as a filter if we use ubiquitous kernels, such as RBF, rational quadratic, arc sine, etc \cite{wilson2013gaussian}. The spectral mixture (SM) kernel should be used for the pattern discovery with GP. However, this kernel is very sensitive to its parameter initialization; the results vary substantially with varying initializations. Also, GPs are generally harder to train on large datasets because of expensive matrix inversion operations. With several underlying approximations and numerical schemes, $\mathtt{GPyTorch}$ is able to make GPs scalable \cite{gardner2018gpytorch}, but then it poses some convergence issues during training, and those issues depend on the dataset. 

One of the many reasons researchers tend to use GP is its inherent ability to give confidence bounds along with predictions, which is important for uncertainty quantification. According to recent studies, GPs are unfortunately {\em not} good uncertainty estimators because they focus on one of the many minima in highly non-convex loss space \cite{wilson2022bayesian}. There are better uncertainty quantification methods, such as deep ensembles \cite{lakshminarayanan2017simple}. In short, GPs are not pattern discovery algorithms unless one uses SM kernel, and they are not even good uncertainty estimators. Hence, neural networks such as LSTM are generally preferred over GP, which can do both jobs relatively well. 

\section{Conclusion}\label{sec:conclusion}

This paper answers questions about model and target variable selection for data-driven CF models. Three data-driven models (GP, Kernel-Ridge, and LSTM), three target variables ($a$, $v$, and $s$), and three datasets (ASTAZERO, NAPOLI, and JIANG) were selected to address these two questions. The recommendations of target variables were compared with those related to classical CF models; it turned out that the best-performing target variables were different for both cases, which signifies the importance of the present study. The best-performing target variables for classical CF models and LSTM are $v$ and $s$, whereas these are $a$ and $v$ for GP and Kernel-Ridge; the reason lies in the respective methods' objective function and vector space. 

The GA (used to optimize classical CF models) and LSTM use the same SSE-based objective function and work in parameter spaces. On the other hand, GP and Kernel Ridge work in function spaces; hence, the target variables are different for these models. In terms of model selection, LSTM should be preferred compared to GP and Kernel-Ridge because it's a pattern discovery algorithm. On the other hand, GP acts as a filter rather than a pattern discovery algorithm unless we use the spectral mixture kernel, but that comes with another set of issues. One reason to prefer GP is its ability to estimate uncertainty, but they are often not even good uncertainty estimators either.  

In the end, analysis of variance showed the robustness of recommended target variables, indicating that the recommendations are applicable irrespective of the dataset under consideration. ASTAZERO was collected from automated vehicles and NAPOLI and JIANG were collected from human-driven vehicles, and each has different combinations of traffic density, speed conditions, and so on. Thus, in addition to model type, our recommendations for target variables should apply to a wide variety of datasets.  

\section*{Acknowledgment}
This work was supported in part by NASA Contract Number 80NSSC22PB011.

\section*{Author Contributions}
\textbf{Muhammad Bilal Shahid}: Writing – original draft, Visualization, Methodology, Conceptualization. \textbf{Cody Fleming}: Supervision, Writing – review \& editing, Conceptualization.

\section*{Declaration of competing interest} 
The authors declare that they have no known competing financial interests or personal relationships that could have appeared to influence the work reported in this paper.

\section*{Data availability}
The data along with the code is available on github. The link is given in section \ref{sec:DetImpGP}.

\bibliographystyle{unsrt}
\bibliography{CFMReferences}

\begin{thebibliography}{10}

\bibitem{pourabdollah2017calibration}
Mitra Pourabdollah, Eric Bj{\"a}rkvik, Florian F{\"u}rer, Bj{\"o}rn Lindenberg, and Klaas Burgdorf.
\newblock Calibration and evaluation of car following models using real-world driving data.
\newblock In {\em 2017 IEEE 20th International conference on intelligent transportation systems (ITSC)}, pages 1--6. IEEE, 2017.

\bibitem{vasconcelos2014calibration}
Lu{\'\i}s Vasconcelos, Lu{\'\i}s Neto, S{\'\i}lvia Santos, Ana~Bastos Silva, and {\'A}lvaro Seco.
\newblock Calibration of the gipps car-following model using trajectory data.
\newblock {\em Transportation research procedia}, 3:952--961, 2014.

\bibitem{markou2019dynamic}
Ioulia Markou, Vasileia Papathanasopoulou, and Constantinos Antoniou.
\newblock Dynamic car--following model calibration using spsa and isres algorithms.
\newblock {\em Periodica Polytechnica Transportation Engineering}, 47(2):146--156, 2019.

\bibitem{wang2021personalized}
Yanbing Wang, Ziran Wang, Kyungtae Han, Prashant Tiwari, and Daniel~B Work.
\newblock Personalized adaptive cruise control via gaussian process regression.
\newblock In {\em 2021 IEEE International Intelligent Transportation Systems Conference (ITSC)}, pages 1496--1502. IEEE, 2021.

\bibitem{soldevila2021car}
Ignasi~Echaniz Soldevila, Victor~L Knoop, and Serge Hoogendoorn.
\newblock Car-following described by blending data-driven and analytical models: a gaussian process regression approach.
\newblock {\em Transportation research record}, 2675(12):1202--1213, 2021.

\bibitem{he2015simple}
Zhengbing He, Liang Zheng, and Wei Guan.
\newblock A simple nonparametric car-following model driven by field data.
\newblock {\em Transportation Research Part B: Methodological}, 80:185--201, 2015.

\bibitem{punzo2012can}
Vincenzo Punzo, Biagio Ciuffo, and Marcello Montanino.
\newblock Can results of car-following model calibration based on trajectory data be trusted?
\newblock {\em Transportation research record}, 2315(1):11--24, 2012.

\bibitem{punzo2021calibration}
Vincenzo Punzo, Zuduo Zheng, and Marcello Montanino.
\newblock About calibration of car-following dynamics of automated and human-driven vehicles: Methodology, guidelines and codes.
\newblock {\em Transportation Research Part C: Emerging Technologies}, 128:103165, 2021.

\bibitem{di2014distributed}
Mario Di~Bernardo, Alessandro Salvi, and Stefania Santini.
\newblock Distributed consensus strategy for platooning of vehicles in the presence of time-varying heterogeneous communication delays.
\newblock {\em IEEE Transactions on Intelligent Transportation Systems}, 16(1):102--112, 2014.

\bibitem{di2019cooperative}
Marco Di~Vaio, Giovanni Fiengo, Alberto Petrillo, Alessandro Salvi, Stefania Santini, and Manuela Tufo.
\newblock Cooperative shock waves mitigation in mixed traffic flow environment.
\newblock {\em IEEE Transactions on Intelligent Transportation Systems}, 20(12):4339--4353, 2019.

\bibitem{zhang2017hierarchical}
Linjun Zhang, Jing Sun, and G{\'a}bor Orosz.
\newblock Hierarchical design of connected cruise control in the presence of information delays and uncertain vehicle dynamics.
\newblock {\em IEEE Transactions on Control Systems Technology}, 26(1):139--150, 2017.

\bibitem{treiber2006delays}
Martin Treiber, Arne Kesting, and Dirk Helbing.
\newblock Delays, inaccuracies and anticipation in microscopic traffic models.
\newblock {\em Physica A: Statistical Mechanics and its Applications}, 360(1):71--88, 2006.

\bibitem{treiber2015comparing}
Martin Treiber and Venkatesan Kanagaraj.
\newblock Comparing numerical integration schemes for time-continuous car-following models.
\newblock {\em Physica A: Statistical Mechanics and its Applications}, 419:183--195, 2015.

\bibitem{pml2Book}
Kevin~P. Murphy.
\newblock {\em Probabilistic Machine Learning: Advanced Topics}.
\newblock MIT Press, 2023.

\bibitem{hochreiter1997long}
Sepp Hochreiter and J{\"u}rgen Schmidhuber.
\newblock Long short-term memory.
\newblock {\em Neural computation}, 9(8):1735--1780, 1997.

\bibitem{makridis2021openacc}
Michail Makridis, Konstantinos Mattas, Aikaterini Anesiadou, and Biagio Ciuffo.
\newblock Openacc. an open database of car-following experiments to study the properties of commercial acc systems.
\newblock {\em Transportation research part C: emerging technologies}, 125:103047, 2021.

\bibitem{jiang2015some}
Rui Jiang, Mao-Bin Hu, HM~Zhang, Zi-You Gao, Bin Jia, and Qing-Song Wu.
\newblock On some experimental features of car-following behavior and how to model them.
\newblock {\em Transportation Research Part B: Methodological}, 80:338--354, 2015.

\bibitem{punzo2005analysis}
Vincenzo Punzo and Fulvio Simonelli.
\newblock Analysis and comparison of microscopic traffic flow models with real traffic microscopic data.
\newblock {\em Transportation Research Record}, 1934(1):53--63, 2005.

\bibitem{brockfeld2003toward}
Elmar Brockfeld, Reinhart~D K{\"u}hne, Alexander Skabardonis, and Peter Wagner.
\newblock Toward benchmarking of microscopic traffic flow models.
\newblock {\em Transportation research record}, 1852(1):124--129, 2003.

\bibitem{fard2019copula}
Mehdi~Rafati Fard and Afshin~Shariat Mohaymany.
\newblock A copula-based estimation of distribution algorithm for calibration of microscopic traffic models.
\newblock {\em Transportation Research Part C: Emerging Technologies}, 98:449--470, 2019.

\bibitem{xu2020statistical}
Tu~Xu and Jorge Laval.
\newblock Statistical inference for two-regime stochastic car-following models.
\newblock {\em Transportation Research Part B: Methodological}, 134:210--228, 2020.

\bibitem{hammit2018evaluation}
Britton~E Hammit, Ali Ghasemzadeh, Rachel~M James, Mohamed~M Ahmed, and Rhonda~Kae Young.
\newblock Evaluation of weather-related freeway car-following behavior using the shrp2 naturalistic driving study database.
\newblock {\em Transportation research part F: traffic psychology and behaviour}, 59:244--259, 2018.

\bibitem{park2019development}
Minju Park, Yeeun Kim, and Hwasoo Yeo.
\newblock Development of an asymmetric car-following model and simulation validation.
\newblock {\em IEEE Transactions on Intelligent Transportation Systems}, 21(8):3513--3524, 2019.

\bibitem{gunter2019model}
George Gunter, Caroline Janssen, William Barbour, Raphael~E Stern, and Daniel~B Work.
\newblock Model-based string stability of adaptive cruise control systems using field data.
\newblock {\em IEEE Transactions on Intelligent Vehicles}, 5(1):90--99, 2019.

\bibitem{li2020car}
Tenglong Li, Dong Ngoduy, Fei Hui, and Xiangmo Zhao.
\newblock A car-following model to assess the impact of v2v messages on traffic dynamics.
\newblock {\em Transportmetrica B: Transport Dynamics}, 8(1):150--165, 2020.

\bibitem{ye2018prediction}
Fei Ye, Peng Hao, Xuewei Qi, Guoyuan Wu, Kanok Boriboonsomsin, and Matthew~J Barth.
\newblock Prediction-based eco-approach and departure at signalized intersections with speed forecasting on preceding vehicles.
\newblock {\em IEEE Transactions on Intelligent Transportation Systems}, 20(4):1378--1389, 2018.

\bibitem{pei2016empirical}
Xin Pei, Yan Pan, Haixin Wang, SC~Wong, and Keechoo Choi.
\newblock Empirical evidence and stability analysis of the linear car-following model with gamma-distributed memory effect.
\newblock {\em Physica A: Statistical Mechanics and its Applications}, 449:311--323, 2016.

\bibitem{xu2018aware}
Donghao Xu, Huijing Zhao, Franck Guillemard, St{\'e}phane Geronimi, and Fran{\c{c}}ois Aioun.
\newblock Aware of scene vehicles—probabilistic modeling of car-following behaviors in real-world traffic.
\newblock {\em IEEE Transactions on Intelligent Transportation Systems}, 20(6):2136--2148, 2018.

\bibitem{hao2016fuzzy}
Haiming Hao, Wanjing Ma, and Hongfeng Xu.
\newblock A fuzzy logic-based multi-agent car-following model.
\newblock {\em Transportation Research Part C: Emerging Technologies}, 69:477--496, 2016.

\bibitem{punzo2016speed}
Vincenzo Punzo and Marcello Montanino.
\newblock Speed or spacing? cumulative variables, and convolution of model errors and time in traffic flow models validation and calibration.
\newblock {\em Transportation Research Part B: Methodological}, 91:21--33, 2016.

\bibitem{hensman2012gpy}
James Hensman, N~Fusi, R~Andrade, N~Durrande, A~Saul, M~Zwiessele, and ND~Lawrence.
\newblock Gpy: A gaussian process framework in python, 2012.

\bibitem{scikit-learn}
F.~Pedregosa, G.~Varoquaux, A.~Gramfort, V.~Michel, B.~Thirion, O.~Grisel, M.~Blondel, P.~Prettenhofer, R.~Weiss, V.~Dubourg, J.~Vanderplas, A.~Passos, D.~Cournapeau, M.~Brucher, M.~Perrot, and E.~Duchesnay.
\newblock Scikit-learn: Machine learning in {P}ython.
\newblock {\em Journal of Machine Learning Research}, 12:2825--2830, 2011.

\bibitem{venkatraman2015improving}
Arun Venkatraman, Martial Hebert, and J~Andrew Bagnell.
\newblock Improving multi-step prediction of learned time series models.
\newblock In {\em Twenty-Ninth AAAI Conference on Artificial Intelligence}, 2015.

\bibitem{ross2011reduction}
St{\'e}phane Ross, Geoffrey Gordon, and Drew Bagnell.
\newblock A reduction of imitation learning and structured prediction to no-regret online learning.
\newblock In {\em Proceedings of the fourteenth international conference on artificial intelligence and statistics}, pages 627--635. JMLR Workshop and Conference Proceedings, 2011.

\bibitem{zhou2017recurrent}
Mofan Zhou, Xiaobo Qu, and Xiaopeng Li.
\newblock A recurrent neural network based microscopic car following model to predict traffic oscillation.
\newblock {\em Transportation research part C: emerging technologies}, 84:245--264, 2017.

\bibitem{huang2018car}
Xiuling Huang, Jie Sun, and Jian Sun.
\newblock A car-following model considering asymmetric driving behavior based on long short-term memory neural networks.
\newblock {\em Transportation research part C: emerging technologies}, 95:346--362, 2018.

\bibitem{cheng2022bayesian}
Zhanhong Cheng, Xudong Wang, Xinyuan Chen, Martin Tr{\'e}panier, and Lijun Sun.
\newblock Bayesian calibration of traffic flow fundamental diagrams using gaussian processes.
\newblock {\em IEEE Open Journal of Intelligent Transportation Systems}, 3:763--771, 2022.

\bibitem{ma2020sequence}
Lijing Ma and Shiru Qu.
\newblock A sequence to sequence learning based car-following model for multi-step predictions considering reaction delay.
\newblock {\em Transportation research part C: emerging technologies}, 120:102785, 2020.

\bibitem{paszke2019pytorch}
Adam Paszke, Sam Gross, Francisco Massa, Adam Lerer, James Bradbury, Gregory Chanan, Trevor Killeen, Zeming Lin, Natalia Gimelshein, Luca Antiga, Alban Desmaison, Andreas Köpf, Edward Yang, Zach DeVito, Martin Raison, Alykhan Tejani, Sasank Chilamkurthy, Benoit Steiner, Lu~Fang, Junjie Bai, and Soumith Chintala.
\newblock Pytorch: An imperative style, high-performance deep learning library, 2019.

\bibitem{wilson2013gaussian}
Andrew~Gordon Wilson and Ryan~Prescott Adams.
\newblock Gaussian process kernels for pattern discovery and extrapolation, 2013.

\bibitem{gardner2018gpytorch}
Jacob~R Gardner, Geoff Pleiss, David Bindel, Kilian~Q Weinberger, and Andrew~Gordon Wilson.
\newblock Gpytorch: Blackbox matrix-matrix gaussian process inference with gpu acceleration.
\newblock In {\em Advances in Neural Information Processing Systems}, 2018.

\bibitem{wilson2022bayesian}
Andrew~Gordon Wilson and Pavel Izmailov.
\newblock Bayesian deep learning and a probabilistic perspective of generalization, 2022.

\bibitem{lakshminarayanan2017simple}
Balaji Lakshminarayanan, Alexander Pritzel, and Charles Blundell.
\newblock Simple and scalable predictive uncertainty estimation using deep ensembles, 2017.

\end{thebibliography}

\pagebreak
\appendix
\counterwithin{figure}{section}

\section{Performances of models}\label{sec:results}
\subsection{Classical CF Models}\label{sec:classical(a & v)}
\begin{figure}[h]
    \begin{minipage}[h]{0.89\linewidth}
        \begin{center}
        \includegraphics[width=1.1\linewidth]{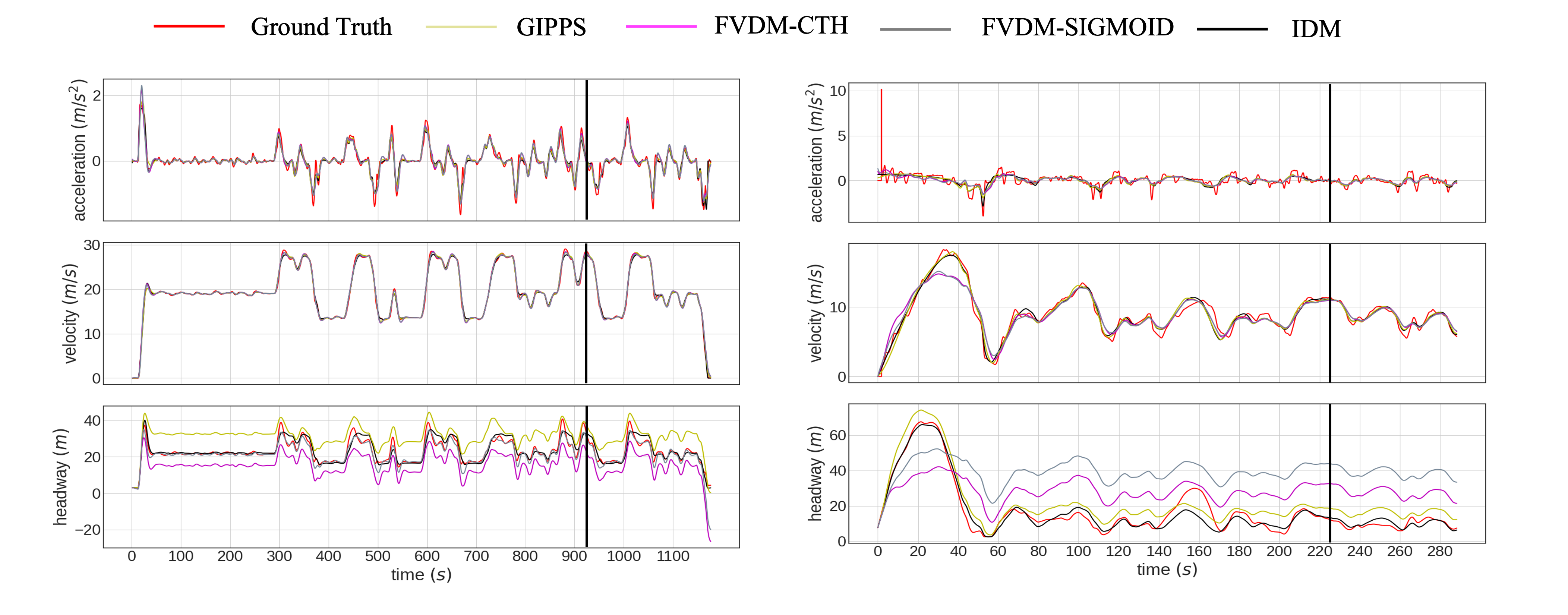}
        \captionsetup{justification=centering, margin=4.4cm,labelformat=empty}
        \vspace*{-10mm}
        \subcaption{\label{fig:classicalASTA(v)}}
        \end{center} 
    \end{minipage}
    \vfill
    \vspace*{-1.3mm}
    \begin{minipage}[h]{0.89\linewidth}
        \begin{center}
        \includegraphics[width=1.1\linewidth]{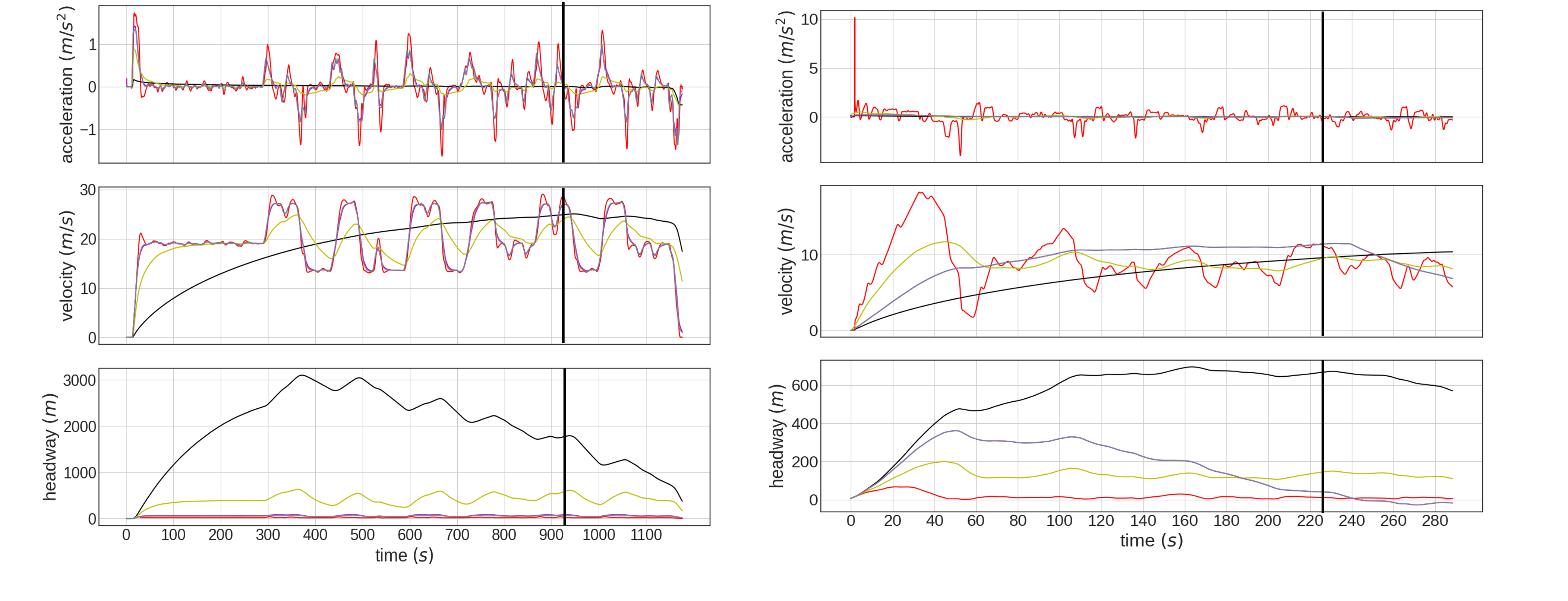}
        \captionsetup{justification=centering, margin=4.4cm, labelformat=empty}
        \vspace*{-11mm}
        \subcaption{\label{fig:classicalASTA(a)}}
        \end{center}
    \end{minipage}
    \vspace*{-3mm}
    \caption{The performance of four classical CF models with their respective learned parameters on
    \textbf{(a)} ASTAZERO ($v$) [Left] \& JIANG ($v$) [Right] \textbf{(b)} ASTAZERO ($a$) [Left] \& JIANG ($a$) [Right]. The black line on each plot indicates the 80/20\% split between training and test data respectively.}
    \label{fig:classical(v)and(a)}
\end{figure}
\clearpage

\subsection{GP \& KRR}
\begin{figure}[h]
    \begin{minipage}[h]{0.47\linewidth}
        \begin{center}
        \includegraphics[width=1.12\linewidth]{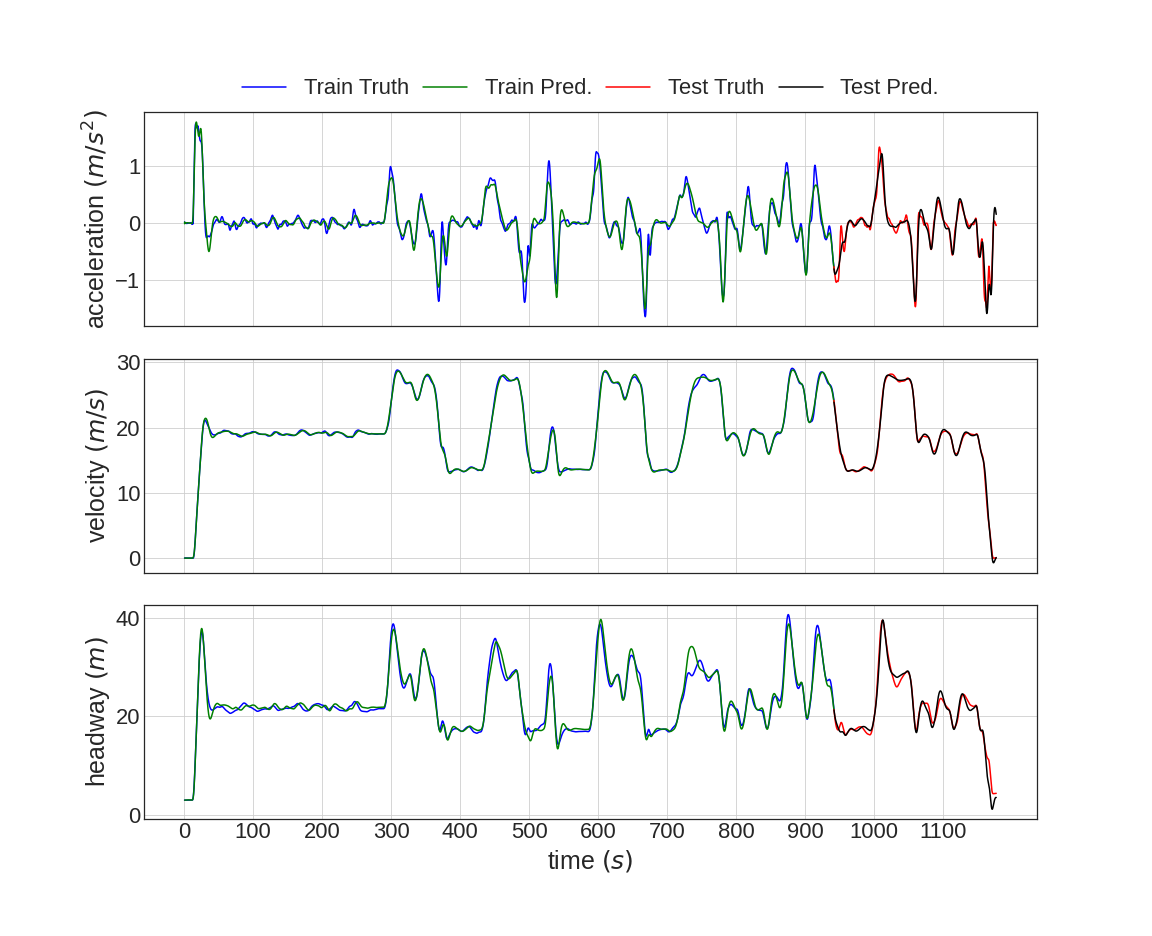}
        \captionsetup{justification=centering, margin=4.4cm, labelformat=empty}
        \vspace*{-11mm}
        \subcaption{\label{fig:GPAsta(v)}}
        \end{center} 
    \end{minipage}
    \begin{minipage}[h]{0.47\linewidth}
        \begin{center}
        \includegraphics[width=1.12\linewidth]{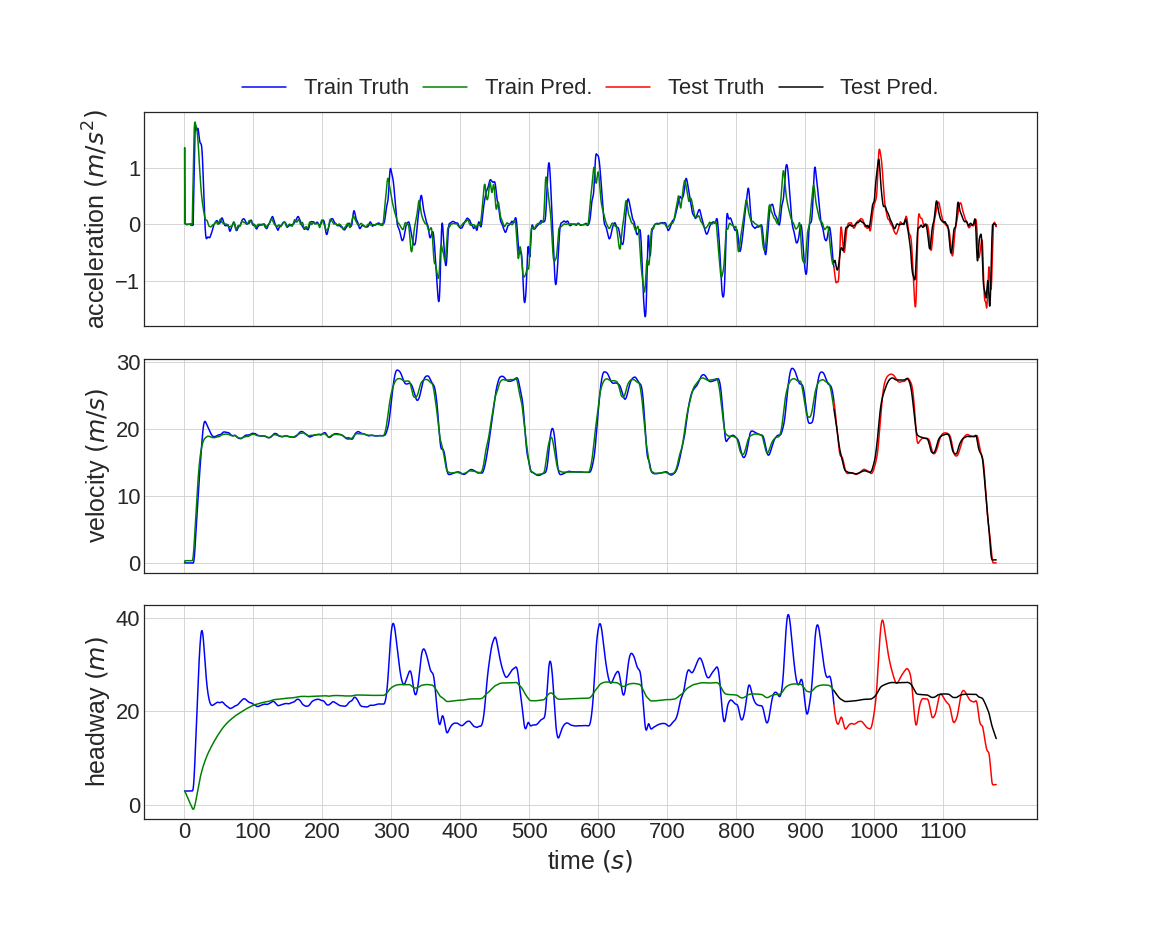}
        \captionsetup{justification=centering, margin=4.4cm, labelformat=empty}
        \vspace*{-11mm}
        \subcaption{\label{fig:KRRAsta(v)}}
        \end{center}
    \end{minipage}
    \vfill
    \vspace*{-1.5mm}
    \begin{minipage}[h]{0.47\linewidth}
        \begin{center}
        \includegraphics[width=1.12\linewidth]{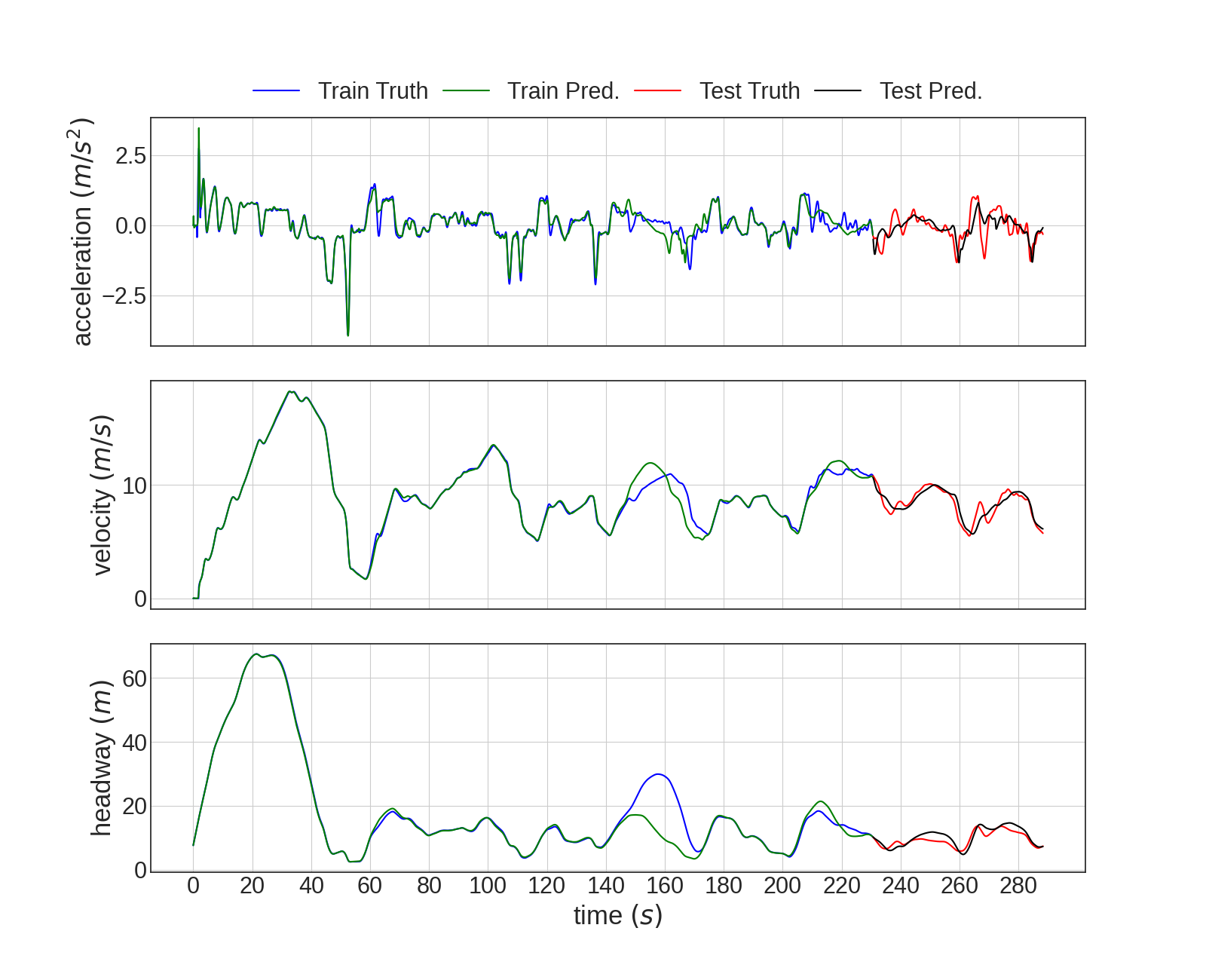}
        \captionsetup{justification=centering, margin=4.4cm, labelformat=empty}
        \vspace*{-11mm}
        \subcaption{\label{fig:GPJiang(v)}}
        \end{center}
    \end{minipage}
    \begin{minipage}[h]{0.47\linewidth}
        \begin{center}
        \includegraphics[width=1.12\linewidth]{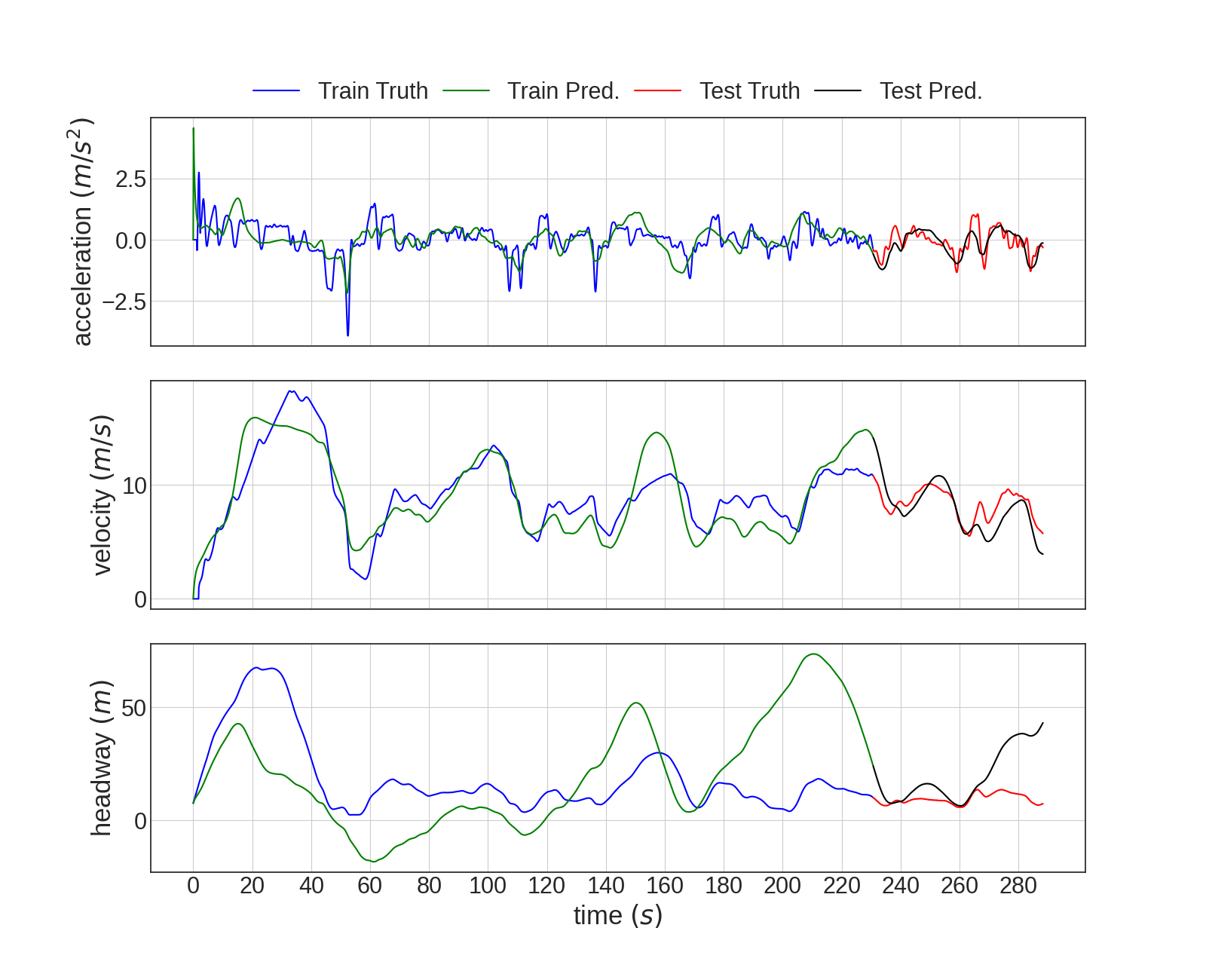} 
        \captionsetup{justification=centering, margin=4.4cm, labelformat=empty}
        \vspace*{-11mm}
        \subcaption{\label{fig:KRRJiang(v)}}
        \end{center}
    \end{minipage}
    \vspace*{-3mm}
    \caption{The performance of GP \& Kernel Ridge on the ASTAZERO and JIANG datasets with $v$ as the target variable (a) GP on ASTAZERO ($v$) (b) KRR on ASTAZERO ($v$) (c) GP on JIANG ($v$) (d) KRR on JIANG ($v$)}
    \label{fig:GPandKRR(v)}
\end{figure}
\clearpage

\begin{figure}[h]
    \begin{minipage}[h]{0.47\linewidth}
        \begin{center}
        \includegraphics[width=1.12\linewidth]{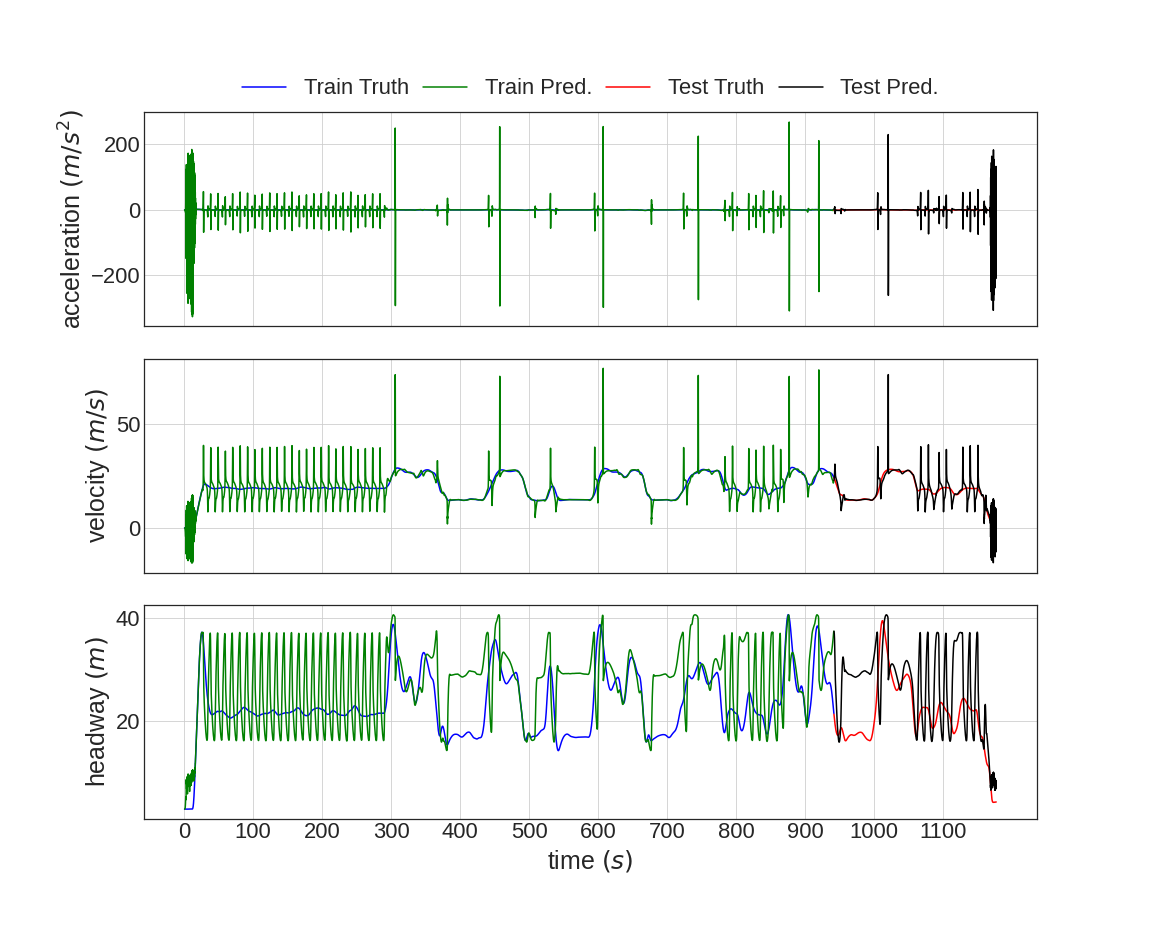}
        \captionsetup{justification=centering, margin=4.4cm, labelformat=empty}
        \vspace*{-11mm}
        \subcaption{\label{fig:GPAsta(s)}}
        \end{center} 
    \end{minipage}
    \begin{minipage}[h]{0.47\linewidth}
        \begin{center}
        \includegraphics[width=1.12\linewidth]{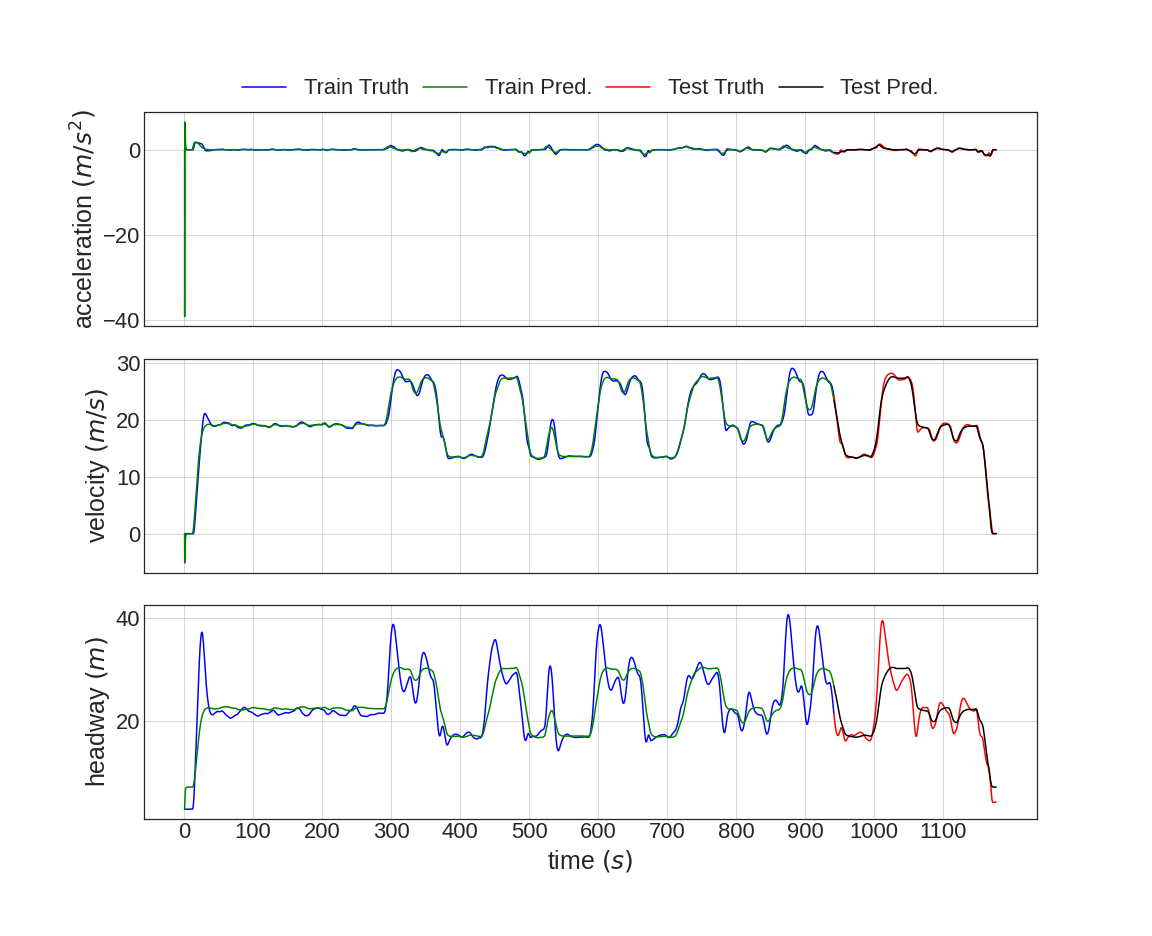}
        \captionsetup{justification=centering, margin=4.4cm, labelformat=empty}
        \vspace*{-11mm}
        \subcaption{\label{fig:KRRAsta(s)}}
        \end{center}
    \end{minipage}
    \vfill
    \vspace*{-1.5mm}
    \begin{minipage}[h]{0.47\linewidth}
        \begin{center}
        \includegraphics[width=1.12\linewidth]{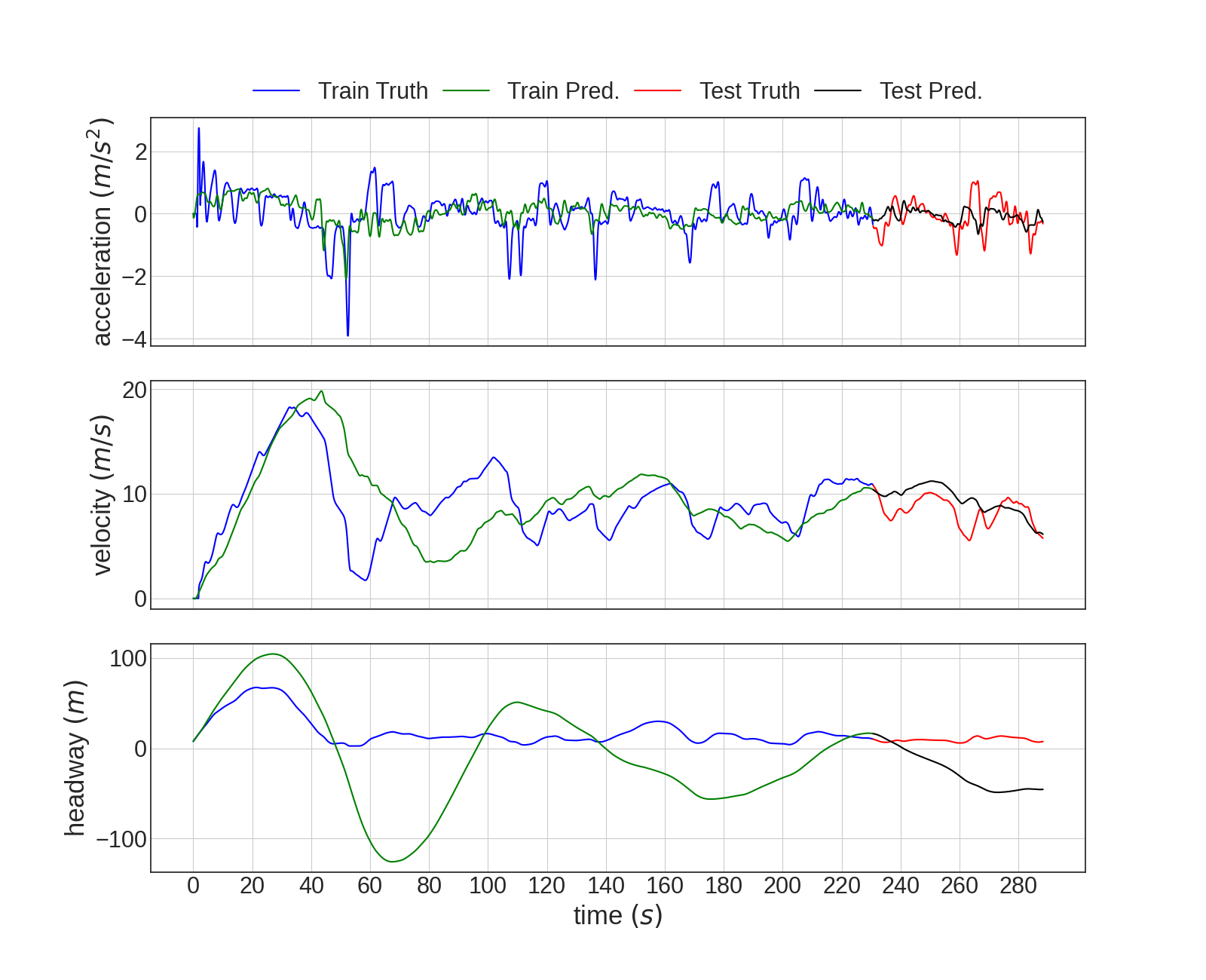}
        \captionsetup{justification=centering, margin=4.4cm, labelformat=empty}
        \vspace*{-11mm}
        \subcaption{\label{fig:GPJiang(s)}}
        \end{center}
    \end{minipage}
    \begin{minipage}[h]{0.47\linewidth}
        \begin{center}
        \includegraphics[width=1.12\linewidth]{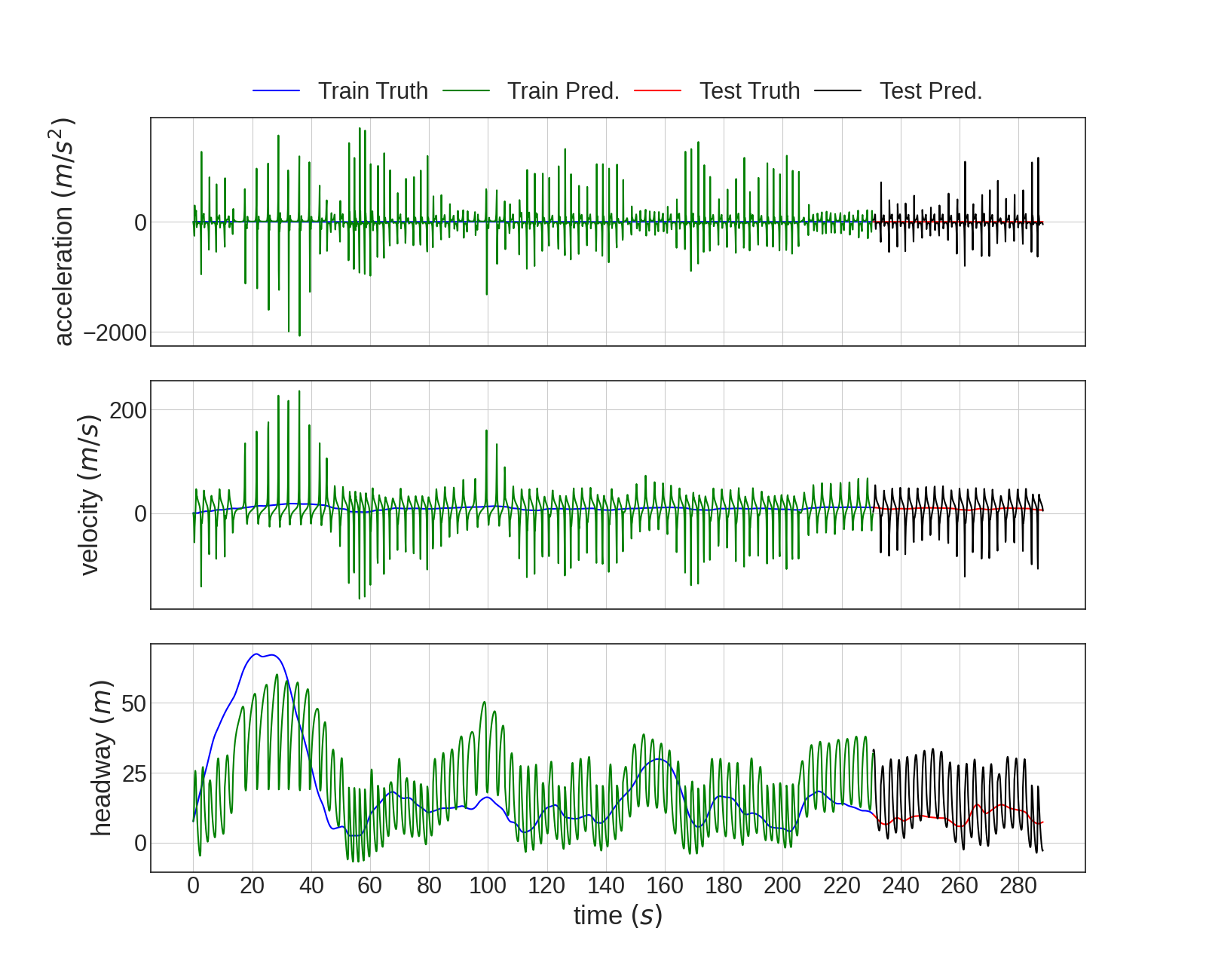} 
        \captionsetup{justification=centering, margin=4.4cm, labelformat=empty}
        \vspace*{-11mm}
        \subcaption{\label{fig:KRRJiang(s)}}
        \end{center}
    \end{minipage}
    \vspace*{-3mm}
    \caption{The performance of GP \& Kernel Ridge on the ASTAZERO and JIANG datasets with $s$ as the target variable (a) GP on ASTAZERO ($s$) (b) KRR on ASTAZERO ($s$) (c) GP on JIANG ($s$) (d) KRR on JIANG ($s$)}
    \label{fig:GPandKRR(s)}
\end{figure}
\clearpage

\subsection{LSTM}
\begin{figure}[h]
    \begin{minipage}[h]{0.47\linewidth}
        \begin{center}
        \includegraphics[width=1.12\linewidth]{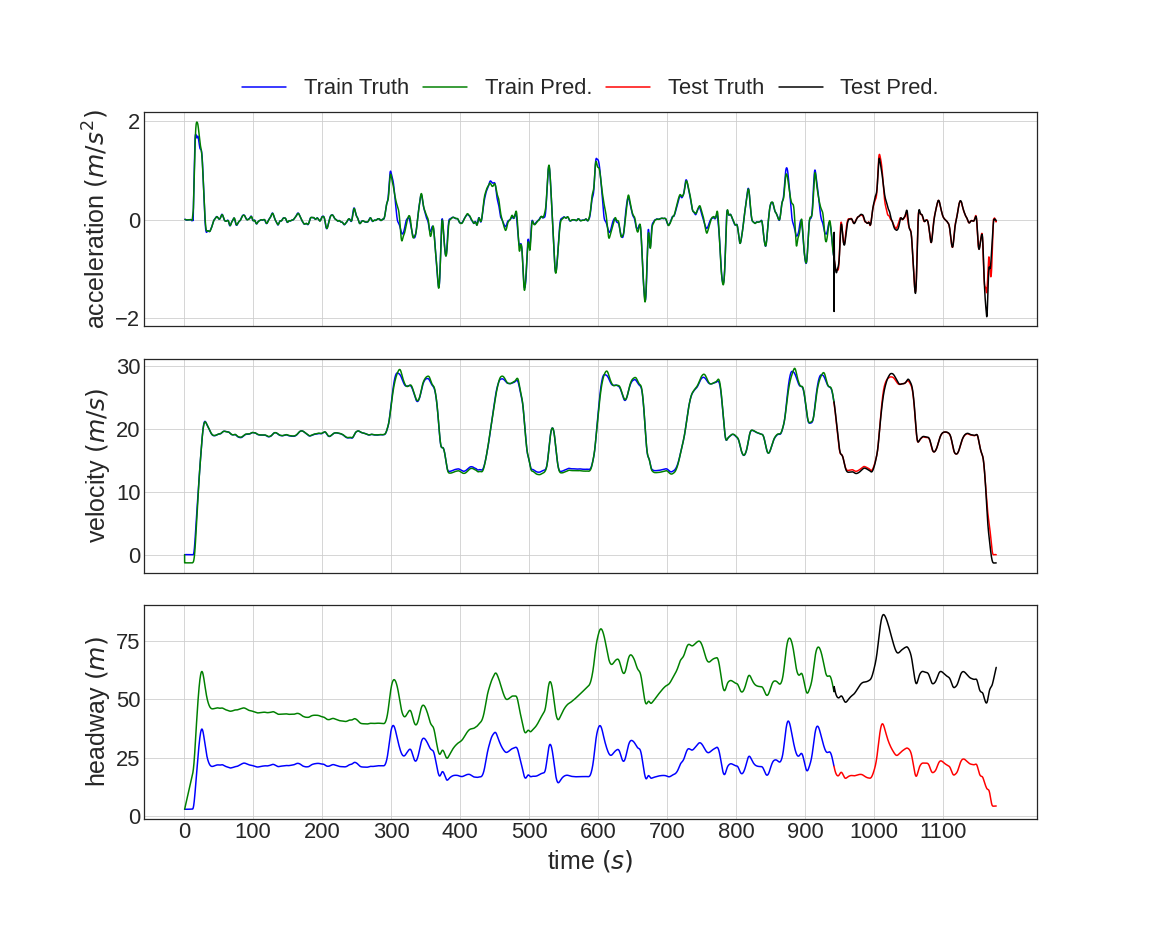}
        \captionsetup{justification=centering, margin=4.4cm, labelformat=empty}
        \vspace*{-11mm}
        \subcaption{\label{fig:LSTMAsta(v)}}
        \end{center} 
    \end{minipage}
    \begin{minipage}[h]{0.47\linewidth}
        \begin{center}
        \includegraphics[width=1.12\linewidth]{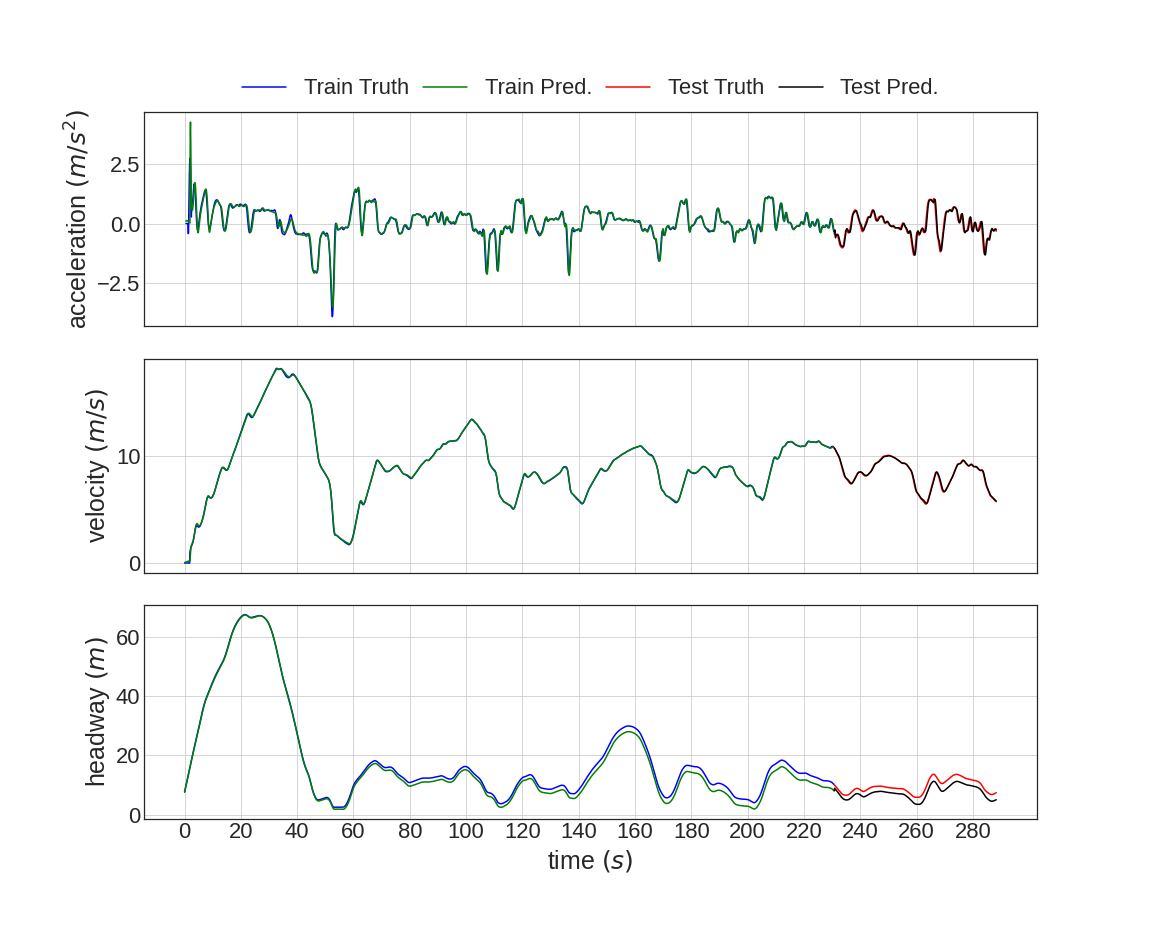}
        \captionsetup{justification=centering, margin=4.4cm, labelformat=empty}
        \vspace*{-11mm}
        \subcaption{\label{fig:LSTMJiang(v)}}
        \end{center}
    \end{minipage}
    \vfill
    \vspace*{-1.5mm}
    \begin{minipage}[h]{0.47\linewidth}
        \begin{center}
        \includegraphics[width=1.12\linewidth]{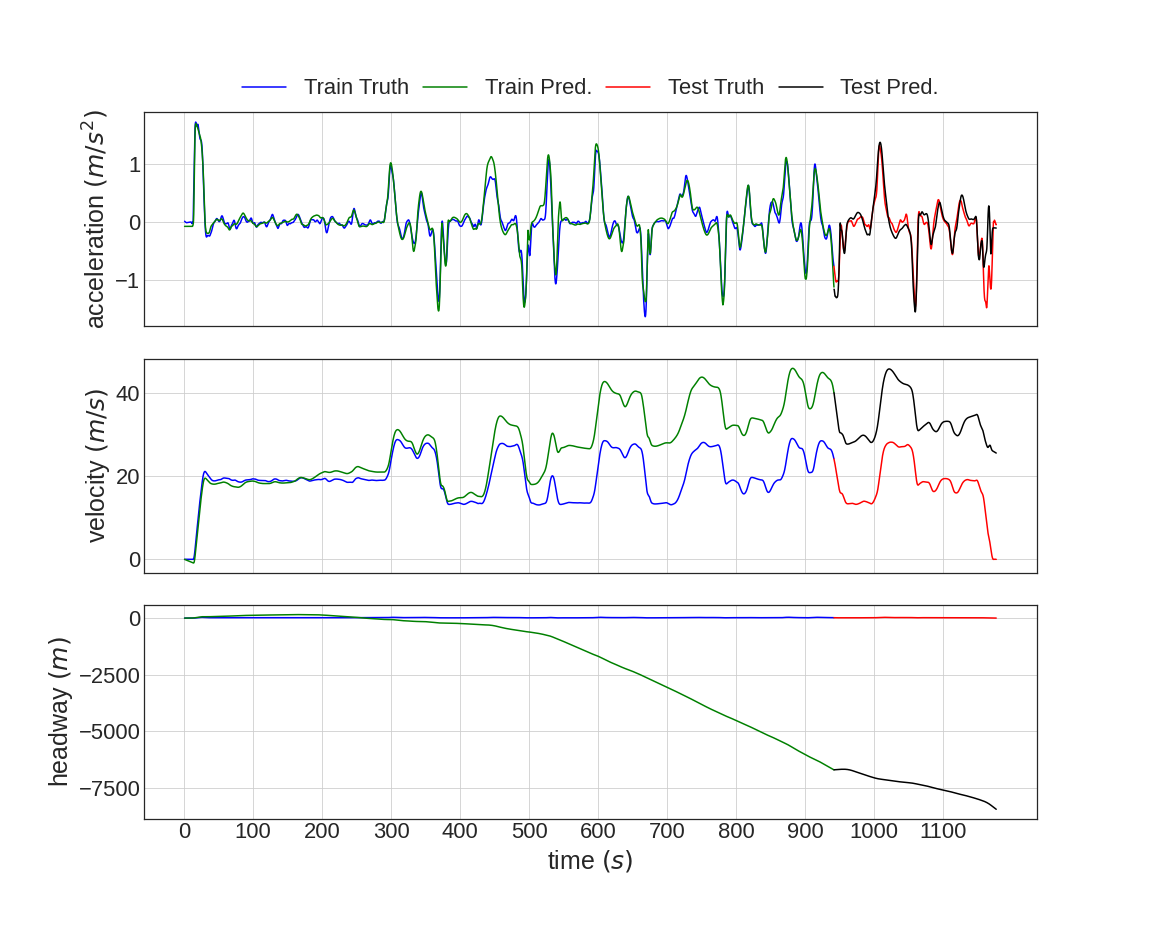}
        \captionsetup{justification=centering, margin=4.4cm, labelformat=empty}
        \vspace*{-11mm}
        \subcaption{\label{fig:LSTMAsta(a)}}
        \end{center}
    \end{minipage}
    \begin{minipage}[h]{0.47\linewidth}
        \begin{center}
        \includegraphics[width=1.12\linewidth]{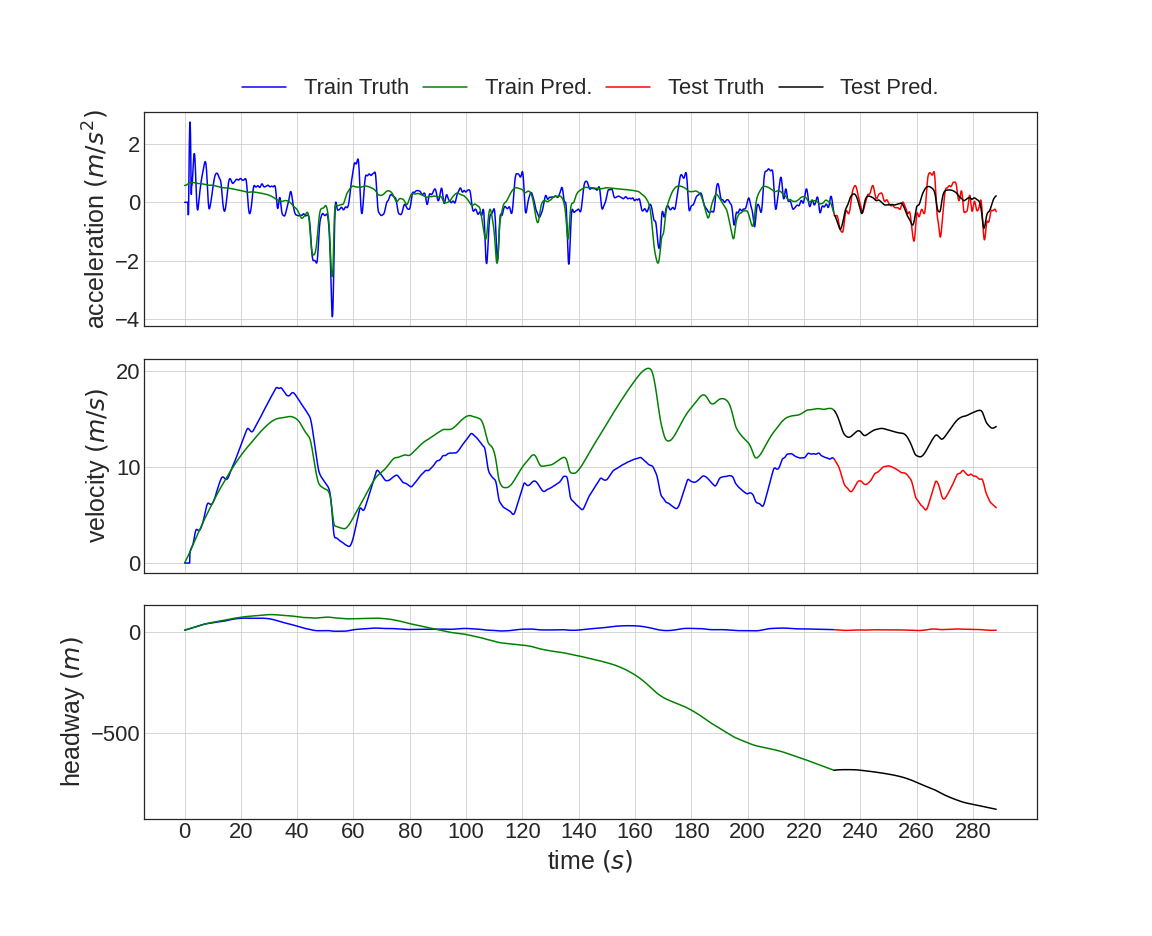} 
        \captionsetup{justification=centering, margin=4.4cm, labelformat=empty}
        \vspace*{-11mm}
        \subcaption{\label{fig:LSTMJiang(a)}}
        \end{center}
    \end{minipage}
    \vspace*{-3mm}
    \caption{The performance of LSTM on (a) ASTAZERO ($v$) (b) JIANG ($v$) (c) ASTAZERO ($a$) (d) JIANG ($a$)}
    \label{fig:LSTM(v)and(a)}
\end{figure}

\clearpage

\subsection{RMSEs across \texorpdfstring{$a$}{a}, \texorpdfstring{$v$}{v}, and \texorpdfstring{$s$}{s} on \texorpdfstring{$\log$}{log} scale}\label{sec:logRMSE}
\begin{figure}[htb!]
  \centering
  \includegraphics[scale=0.55]{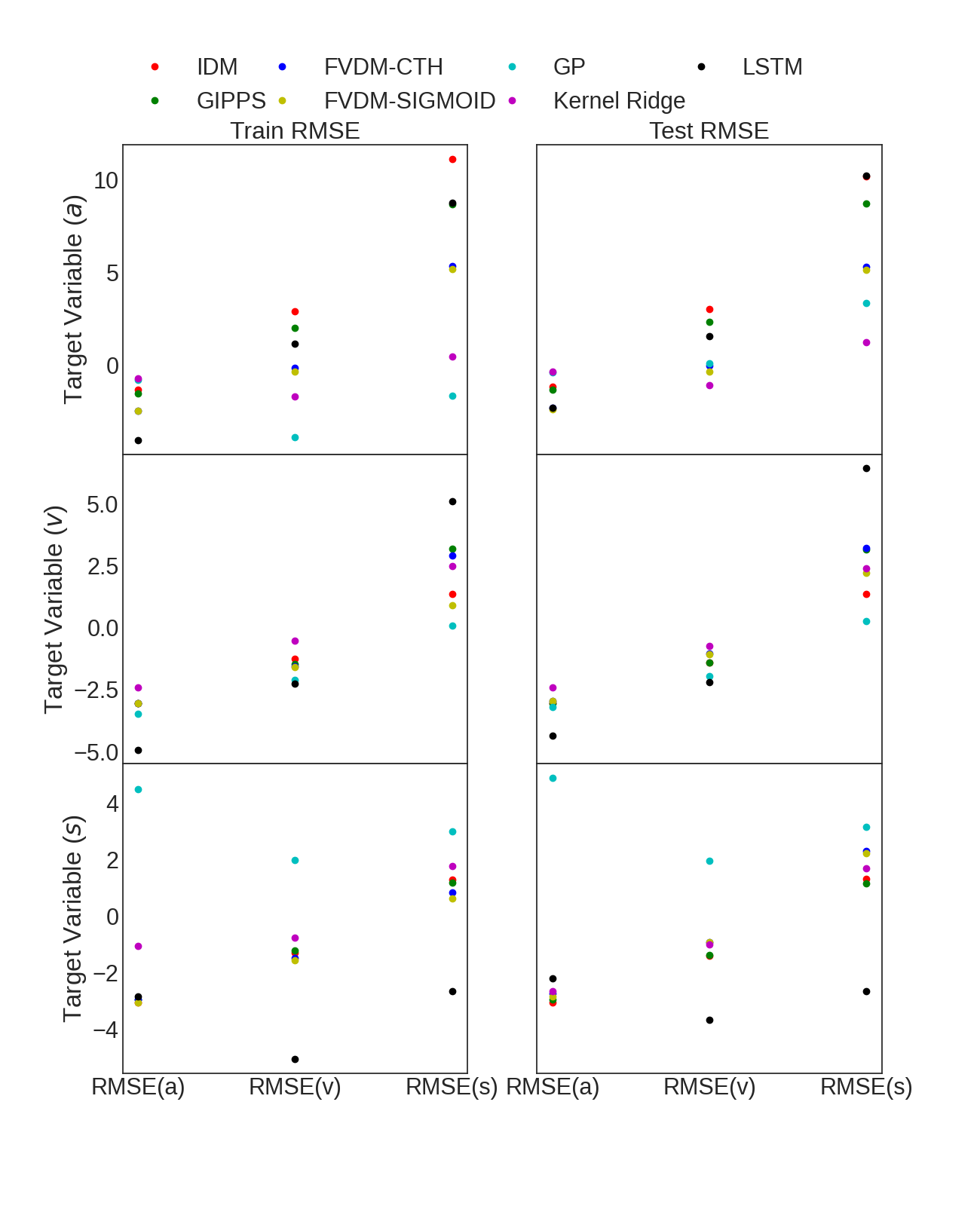}
  \caption{Log of RMSEs of ASTAZERO dataset with all target variables across $a$, $v$, and $s$}
  \label{fig:astaLogRMSE}
\end{figure}

\clearpage

\begin{figure}[htb!]
  \centering
  \includegraphics[scale=0.55]{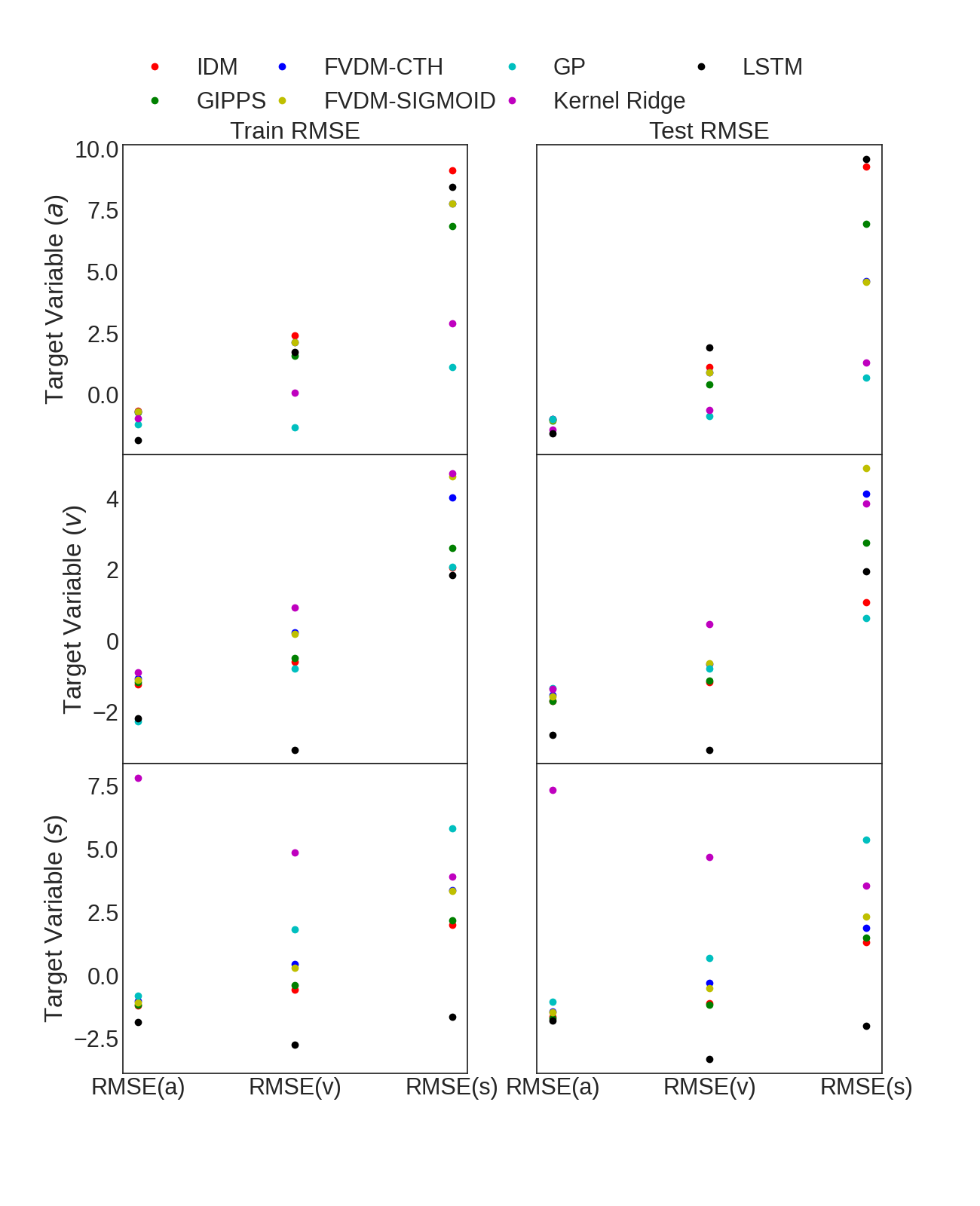}
  \caption{Log of RMSEs of JIANG dataset with all target variables across $a$, $v$, and $s$}
  \label{fig:jiangLogRMSE}
\end{figure}

\clearpage
\begin{figure}[htb!]
  \centering
  \includegraphics[scale=0.55]{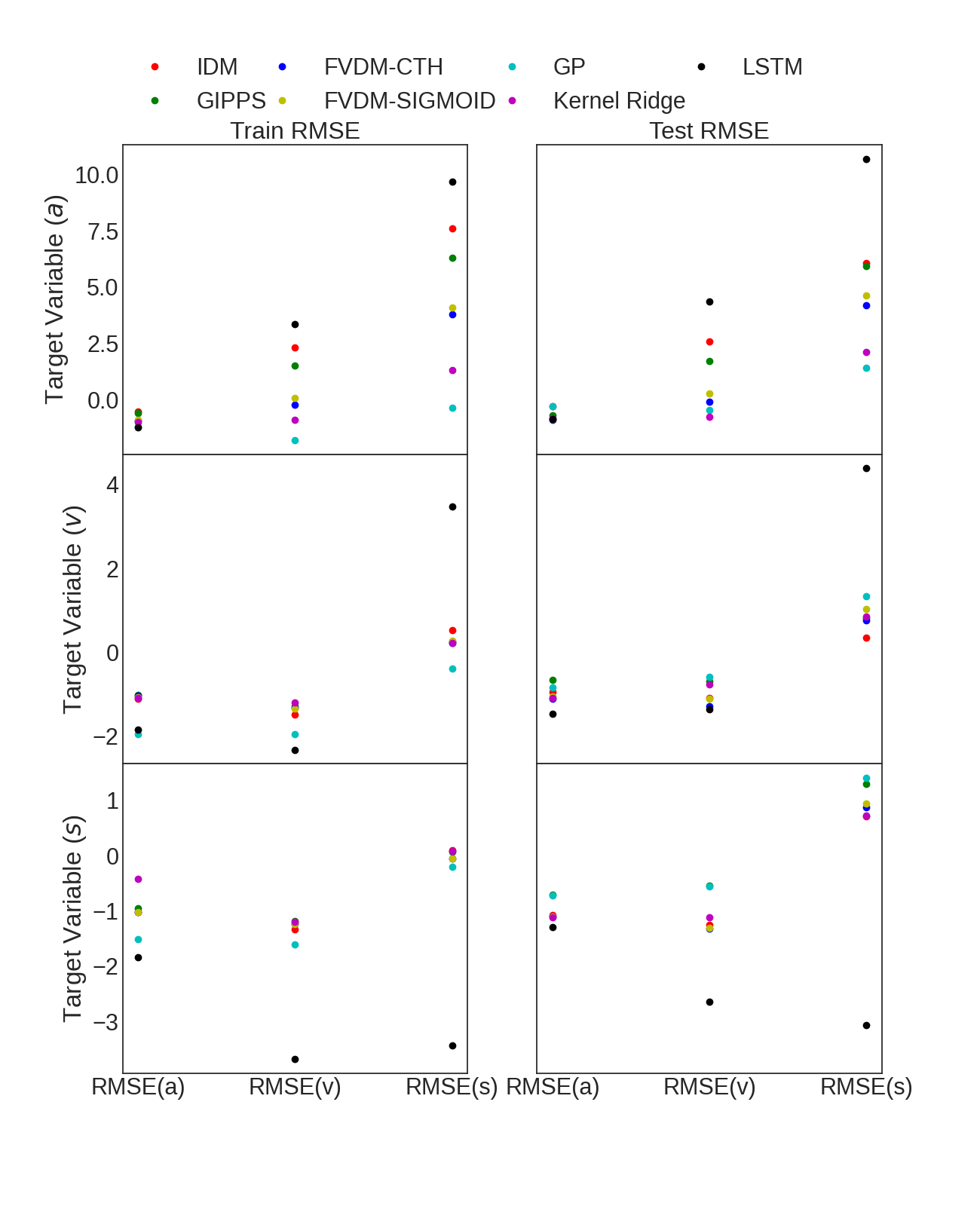}
  \caption{Log of RMSEs of NAPOLI dataset with all target variables across $a$, $v$, and $s$}
  \label{fig:napoliLogRMSE}
\end{figure}

\clearpage
\section{RMSEs across \texorpdfstring{$a$}{a}, \texorpdfstring{$v$}{v}, and \texorpdfstring{$s$}{s} for all combinations}\label{sec:rmse}
The first row of the table below means that the IDM model using dataset ASTAZERO when optimized on $a$ resulted in RMSE of 0.45 on acceleration prediction.\\

\begin{tabular}{|c|c|c|c|}
\hline
RMSE(a)   & Dataset & Model         & Target\\
\hline
0.45   & ASTA & IDM          & a \\
\hline
0.12   & ASTA & IDM          & v \\
\hline
0.12   & ASTA & IDM          & s \\
\hline
0.4    & ASTA & GIPPS        & a \\
\hline
0.12   & ASTA & GIPPS        & v \\
\hline
0.13   & ASTA & GIPPS        & s \\
\hline
0.2    & ASTA & FVDM-CTH     & a \\
\hline
0.13   & ASTA & FVDM-CTH     & v \\
\hline
0.15   & ASTA & FVDM-CTH     & s \\
\hline
0.19   & ASTA & FVDM-SIG & a \\
\hline
0.13   & ASTA & FVDM-SIG & v \\
\hline
0.14   & ASTA & FVDM-SIG & s \\
\hline
0.76   & ASTA & GP           & a \\
\hline
0.11   & ASTA & GP           & v \\
\hline
29.96  & ASTA & GP           & s \\
\hline
0.78   & ASTA & KRR          & a \\
\hline
0.19   & ASTA & KRR          & v \\
\hline
0.16   & ASTA & KRR          & s \\
\hline
0.31   & ASTA & LSTM         & a \\
\hline
0.049  & ASTA & LSTM         & v \\
\hline
0.22   & ASTA & LSTM         & s \\
\hline
0.51   & JIANG    & IDM          & a \\
\hline
0.31   & JIANG    & IDM          & v \\
\hline
0.32   & JIANG    & IDM          & s \\
\hline
0.49   & JIANG    & GIPPS        & a \\
\hline
0.31   & JIANG    & GIPPS        & v \\
\hline
0.31   & JIANG    & GIPPS        & s \\
\hline
0.51   & JIANG    & FVDM-CTH     & a \\
\hline
0.35   & JIANG    & FVDM-CTH     & v \\
\hline
0.37   & JIANG    & FVDM-CTH     & s \\
\hline
0.5    & JIANG    & FVDM-SIGM & a \\
\hline
0.34   & JIANG    & FVDM-SIG & v \\
\hline
0.36   & JIANG    & FVDM-SIG & s \\
\hline
0.51   & JIANG    & GP           & a \\
\hline
0.4    & JIANG    & GP           & v \\
\hline
0.49   & JIANG    & GP           & s \\
\hline
0.38   & JIANG    & KRR          & a \\
\hline
0.39   & JIANG    & KRR          & v \\
\hline
161.67 & JIANG    & KRR          & s \\
\hline
0.34   & JIANG    & LSTM         & a \\
\hline
0.16   & JIANG    & LSTM         & v \\
\hline
0.29   & JIANG    & LSTM         & s \\
\hline
\end{tabular}
\quad
\begin{tabular}{|c|c|c|c|}
\hline
RMSE(a)   & Dataset & Model         & Target\\
\hline
0.82   & NAPOLI   & IDM          & a \\
\hline
0.52   & NAPOLI   & IDM          & v \\
\hline
0.48   & NAPOLI   & IDM          & s \\
\hline
0.62   & NAPOLI   & GIPPS        & a \\
\hline
0.63   & NAPOLI   & GIPPS        & v \\
\hline
0.62   & NAPOLI   & GIPPS        & s \\
\hline
0.54   & NAPOLI   & FVDM-CTH     & a \\
\hline
0.47   & NAPOLI   & FVDM-CTH     & v \\
\hline
0.47   & NAPOLI   & FVDM-CTH     & s \\
\hline
0.55   & NAPOLI   & FVDM-SIG & a \\
\hline
0.48   & NAPOLI   & FVDM-SIG & v \\
\hline
0.47   & NAPOLI   & FVDM-SIG & s \\
\hline
0.82   & NAPOLI   & GP           & a \\
\hline
0.56   & NAPOLI   & GP           & v \\
\hline
0.61   & NAPOLI   & GP           & s \\
\hline
0.56   & NAPOLI   & KRR          & a \\
\hline
0.47   & NAPOLI   & KRR          & v \\
\hline
0.46   & NAPOLI   & KRR          & s \\
\hline
0.56   & NAPOLI   & LSTM         & a \\
\hline
0.36   & NAPOLI   & LSTM         & v \\
\hline
0.41   & NAPOLI   & LSTM         & s \\
\hline
\end{tabular}

\begin{tabular}{|c|c|c|c|}
\hline
RMSE(v)   & Dataset & Model         & Target\\
\hline
8.12   & ASTA & IDM          & a \\
\hline
0.38   & ASTA & IDM          & v \\
\hline
0.38   & ASTA & IDM          & s \\
\hline
5.08   & ASTA & GIPPS        & a \\
\hline
0.38   & ASTA & GIPPS        & v \\
\hline
0.39   & ASTA & GIPPS        & s \\
\hline
0.98   & ASTA & FVDM-CTH     & a \\
\hline
0.49   & ASTA & FVDM-CTH     & v \\
\hline
0.53   & ASTA & FVDM-CTH     & s \\
\hline
0.78   & ASTA & FVDM-SIG & a \\
\hline
0.48   & ASTA & FVDM-SIG & v \\
\hline
0.53   & ASTA & FVDM-SIG & s \\
\hline
1.07   & ASTA & GP           & a \\
\hline
0.26   & ASTA & GP           & v \\
\hline
3.93   & ASTA & GP           & s \\
\hline
0.47   & ASTA & KRR          & a \\
\hline
0.6    & ASTA & KRR          & v \\
\hline
0.5    & ASTA & KRR          & s \\
\hline
104.73 & ASTA & LSTM         & a \\
\hline
0.22   & ASTA & LSTM         & v \\
\hline
0.079  & ASTA & LSTM         & s \\
\hline
2.19   & JIANG    & IDM          & a \\
\hline
0.45   & JIANG    & IDM          & v \\
\hline
0.47   & JIANG    & IDM          & s \\
\hline
1.34   & JIANG    & GIPPS        & a \\
\hline
0.46   & JIANG    & GIPPS        & v \\
\hline
0.45   & JIANG    & GIPPS        & s \\
\hline
1.87   & JIANG    & FVDM-CTH     & a \\
\hline
0.64   & JIANG    & FVDM-CTH     & v \\
\hline
0.81   & JIANG    & FVDM-CTH     & s \\
\hline
1.87   & JIANG    & FVDM-SIGM & a \\
\hline
0.65   & JIANG    & FVDM-SIG & v \\
\hline
0.7    & JIANG    & FVDM-SIG & s \\
\hline
0.55   & JIANG    & GP           & a \\
\hline
0.58   & JIANG    & GP           & v \\
\hline
1.62   & JIANG    & GP           & s \\
\hline
0.65   & JIANG    & KRR          & a \\
\hline
1.38   & JIANG    & KRR          & v \\
\hline
25.92  & JIANG    & KRR          & s \\
\hline
3.81   & JIANG    & LSTM         & a \\
\hline
0.12   & JIANG    & LSTM         & v \\
\hline
0.1    & JIANG    & LSTM         & s \\
\hline
\end{tabular}
\quad
\begin{tabular}{|c|c|c|c|}
\hline
RMSE(v)   & Dataset & Model         & Target\\
\hline
5.98   & NAPOLI   & IDM          & a \\
\hline
0.47   & NAPOLI   & IDM          & v \\
\hline
0.42   & NAPOLI   & IDM          & s \\
\hline
3.31   & NAPOLI   & GIPPS        & a \\
\hline
0.618  & NAPOLI   & GIPPS        & v \\
\hline
0.68   & NAPOLI   & GIPPS        & s \\
\hline
0.94   & NAPOLI   & FVDM-CTH     & a \\
\hline
0.41   & NAPOLI   & FVDM-CTH     & v \\
\hline
0.4    & NAPOLI   & FVDM-CTH     & s \\
\hline
1.21   & NAPOLI   & FVDM-SIG & a \\
\hline
0.46   & NAPOLI   & FVDM-SIG & v \\
\hline
0.4    & NAPOLI   & FVDM-SIG & s \\
\hline
0.73   & NAPOLI   & GP           & a \\
\hline
0.67   & NAPOLI   & GP           & v \\
\hline
0.68   & NAPOLI   & GP           & s \\
\hline
0.59   & NAPOLI   & KRR          & a \\
\hline
0.58   & NAPOLI   & KRR          & v \\
\hline
0.46   & NAPOLI   & KRR          & s \\
\hline
20.47  & NAPOLI   & LSTM         & a \\
\hline
0.39   & NAPOLI   & LSTM         & v \\
\hline
0.16   & NAPOLI   & LSTM         & s \\
\hline
\end{tabular}

\begin{tabular}{|c|c|c|c|}
\hline
RMSE(s)   & Dataset & Model         & Target\\
\hline
1152.92  & ASTA   & IDM      & a \\
\hline
2.57     & ASTA   & IDM      & v \\
\hline
2.52     & ASTA   & IDM      & s \\
\hline
418.31   & ASTA   & GIPPS    & a \\
\hline
8.92     & ASTA   & GIPPS    & v \\
\hline
2.24     & ASTA   & GIPPS    & s \\
\hline
39.29    & ASTA   & FVDM-CTH & a \\
\hline
9.2      & ASTA   & FVDM-CTH & v \\
\hline
4.96     & ASTA   & FVDM-CTH & s \\
\hline
35.11    & ASTA   & FVDM-SIG & a \\
\hline
4.59     & ASTA   & FVDM-SIG & v \\
\hline
4.73     & ASTA   & FVDM-SIG & s \\
\hline
10.13    & ASTA   & GP       & a \\
\hline
1.19     & ASTA   & GP       & v \\
\hline
9.01     & ASTA   & GP       & s \\
\hline
2.36     & ASTA   & KRR      & a \\
\hline
5.2      & ASTA   & KRR      & v \\
\hline
3.26     & ASTA   & KRR      & s \\
\hline
47918.08 & ASTA   & LSTM     & a \\
\hline
85.88    & ASTA   & LSTM     & v \\
\hline
0.16     & ASTA   & LSTM     & s \\
\hline
622.53   & JIANG  & IDM      & a \\
\hline
2.12     & JIANG  & IDM      & v \\
\hline
2.51     & JIANG  & IDM      & s \\
\hline
122.23   & JIANG  & GIPPS    & a \\
\hline
6.81     & JIANG  & GIPPS    & v \\
\hline
2.81     & JIANG  & GIPPS    & s \\
\hline
24.56    & JIANG  & FVDM-CTH & a \\
\hline
17.68    & JIANG  & FVDM-CTH & v \\
\hline
3.66     & JIANG  & FVDM-CTH & s \\
\hline
24.18    & JIANG  & FVDM-SIG & a \\
\hline
28.92    & JIANG  & FVDM-SIG & v \\
\hline
5.05     & JIANG  & FVDM-SIG & s \\
\hline
1.62     & JIANG  & GP       & a \\
\hline
1.56     & JIANG  & GP       & v \\
\hline
41.31    & JIANG  & GP       & s \\
\hline
2.5      & JIANG  & KRR      & a \\
\hline
14.55    & JIANG  & KRR      & v \\
\hline
11.71    & JIANG  & KRR      & s \\
\hline
770.2    & JIANG  & LSTM     & a \\
\hline
3.88     & JIANG  & LSTM     & v \\
\hline
0.25     & JIANG  & LSTM     & s \\
\hline
\end{tabular}
\quad
\begin{tabular}{|c|c|c|c|}
\hline
RMSE(s)   & Dataset & Model         & Target\\
\hline
67.42    & NAPOLI & IDM      & a \\
\hline
1.28     & NAPOLI & IDM      & v \\
\hline
1.63     & NAPOLI & IDM      & s \\
\hline
61.18    & NAPOLI & GIPPS    & a \\
\hline
1.77     & NAPOLI & GIPPS    & v \\
\hline
2.44     & NAPOLI & GIPPS    & s \\
\hline
18.45    & NAPOLI & FVDM-CTH & a \\
\hline
1.7      & NAPOLI & FVDM-CTH & v \\
\hline
1.83     & NAPOLI & FVDM-CTH & s \\
\hline
24.67    & NAPOLI & FVDM-SIG & a \\
\hline
2.05     & NAPOLI & FVDM-SIG & v \\
\hline
1.92     & NAPOLI & FVDM-SIG & s \\
\hline
2.68     & NAPOLI & GP       & a \\
\hline
2.54     & NAPOLI & GP       & v \\
\hline
2.65     & NAPOLI & GP       & s \\
\hline
4.41     & NAPOLI & KRR      & a \\
\hline
1.8      & NAPOLI & KRR      & v \\
\hline
1.66     & NAPOLI & KRR      & s \\
\hline
1654.78     & NAPOLI & LSTM     & a \\
\hline
20.88    & NAPOLI & LSTM     & v \\
\hline
0.12  & NAPOLI & LSTM     & s \\
\hline
\end{tabular}

\section{Kernels}
The mathematical forms of different kernels used in this study are given below. For interested readers, further information about each kernel can be found in the documentation of GPy \cite{hensman2012gpy}.
\begin{align*}
    \textbf{RBF}\quad \quad \quad \quad \quad \quad k(r) &= \sigma^{2}\text{exp}\left(-\frac{1}{2}r^{2}\right) \quad \text{where} \quad r = \sqrt{\sum^{\text{input\_dim}}_{i=1}\frac{\norm{x_{i}-y_{i}}^{2}}{l^{2}_{i}}} \\ 
    \textbf{Exponential}\quad \quad \quad \quad \quad \quad k(r) &= \sigma^{2}\text{exp}\left(-r\right)\\
    \textbf{Rational Quadratic}\quad \quad \quad \quad \quad \quad k(r) &= \sigma^{2}\left(1+\frac{r^{2}}{2}\right)^{-\alpha}\\
    \textbf{MLP}\quad \quad \quad \quad \quad \quad k(x,y) &= \sigma^{2}\frac{2}{\pi}\text{asin}\left(\frac{\sigma^{2}_{w}x^{T}y+\sigma^{2}_{b}}{\sqrt{\sigma^{2}_{w}x^{T}x+\sigma^{2}_{b}+1}\sqrt{\sigma^{2}_{w}y^{T}y+\sigma^{2}_{b}+1}}\right)\\
    \textbf{Matérn 3/2}\quad \quad \quad \quad \quad \quad k(r) &= \sigma^{2}\left(1+\sqrt{3}r\right)\text{exp}\left(-\sqrt{3}r\right) \\
    \textbf{Matérn 5/2}\quad \quad \quad \quad \quad \quad k(r) &= \sigma^{2}\left(1+\sqrt{5}r+\frac{5}{3}r^{2}\right)\text{exp}\left(-\sqrt{5}r\right)
\end{align*}

\end{document}